\documentclass[ijoc,sgblindrev]{informs4}
\usepackage{txfonts}
\usepackage{bbm}

\DeclareMathVersion{bold}

\usepackage[ruled,linesnumbered,noend]{algorithm2e}
\usepackage{caption}
\usepackage{color}
\usepackage{subfigure}
\usepackage{appendix}
\usepackage{amsmath, amsfonts}
\usepackage{bm}

\OneAndAHalfSpacedXI

\usepackage{natbib}
 \bibpunct[, ]{(}{)}{,}{a}{}{,}%
 \def\bibfont{\footnotesize}%

\usepackage[bookmarks=false]{hyperref}
\hypersetup{
   colorlinks=true,
   linkcolor=[RGB]{17, 14, 178},
   citecolor= [RGB]{17, 14, 178},
   filecolor= [RGB]{17, 14, 178},
   urlcolor= [RGB]{17, 14, 178},
	pdfpagemode=FullScreen,
    hypertexnames=false
   }

\newcommand{\E}{{\rm E}}

\TheoremsNumberedThrough
\ECRepeatTheorems
\EquationsNumberedBySection

\begin{document}

\newcommand{\papertitle}{Sequential Additivity in Distributionally Robust Ranking and Selection}

\RUNTITLE{\papertitle}

\TITLE{\large \papertitle}
\ABSTRACT{
Ranking and selection (R\&S) seeks to identify the alternative with the best mean performance from a finite collection of simulated alternatives. Its practical value depends on accurate simulation input modeling, which is often hindered by input uncertainty arising from limited data. Distributionally robust R\&S (DRR\&S) addresses this challenge by considering several plausible input distributions and selecting the alternative with the best worst-case mean performance, resulting in a multiplicative number of scenarios. Existing static and oracle analyses suggest that efficient sampling should instead be additive, concentrating on only a small number of critical scenarios. We introduce sequential additivity, which characterizes how this structure emerges from adaptive sequential procedures. We first establish an algorithm-independent sampling lower bound: any consistent DRR\&S procedure must sample at least this additive number of scenarios infinitely often. We then study a simplified additive allocation (AA) procedure abstracted from practical sequential designs.
Using boundary-crossing arguments, we derive a finite-budget upper bound on its probability of incorrect selection and show that this probability decays exponentially as the budget grows. Moreover, AA attains the necessary sampling lower bound exactly, showing that additivity can be achieved in the strongest possible sense. Surprisingly, the scenarios sampled infinitely often need not be the true worst-case scenarios, showing that worst-case scenario identification may not be necessary for sufficient exploration in DRR\&S. To generalize these insights, we introduce a general additive allocation (GAA) framework that incorporates sampling rules from traditional R\&S in a modular fashion. Under suitable exploration conditions, GAA procedures retain the key properties of AA. Our results lead to a new class of efficient DRR\&S procedures and may deepen the structural understanding of robust R\&S under input uncertainty.

}%

\RUNAUTHOR{Li, Wan, and Hong}
\ARTICLEAUTHORS{%
\AUTHOR{Zaile Li\textsuperscript{a},
Yuchen Wan\textsuperscript{b},
L. Jeff Hong\textsuperscript{c}}
\AFF{\textsuperscript{a} Department of Industrial Systems Engineering and Management, National University of Singapore, Singapore;
\textsuperscript{b} School of Data Science, Fudan University, Shanghai, China; 
\textsuperscript{c} Department of Industrial and Systems Engineering, University of Minnesota, Minneapolis, Minnesota\\
}
\AFF{\textbf{Contact:} \EMAIL{zaile.li@nus.edu.sg} (ZL); \EMAIL{ycwan22@m.fudan.edu.cn} (YW); \EMAIL{lhong@umn.edu} (LJH)}
\vspace{-5pt}
} 

\maketitle

\vspace{-10pt}

\section{Introduction}
\label{sec: intro}

Ranking and selection (R\&S) seeks to identify the best among a finite set of simulated alternatives. Since the seminal work of \citet{bechhofer1954single}, R\&S has been studied extensively under a variety of formulations and methodological approaches; see \citet{hong2021review} for a recent review. In this paper, we focus on the fixed-budget setting, in which the total sampling budget is pre-specified, and the objective is to maximize the probability of correct selection (PCS). A fundamental requirement for a fixed-budget procedure is consistency, meaning that its PCS converges to one as the sampling budget grows. In conventional R\&S, achieving consistency requires every alternative to be sampled infinitely often \citep{hong2021review}; 
stronger properties, such as exponential decay of the probability of incorrect selection (PICS) \citep[e.g.,][]{wu2018analyzing,li2023convergence} and asymptotically optimal budget allocation ratios \citep[e.g.,][]{glynn2004large,gao2017new, peng2018ranking, chen2022balancing}, are also of central interest.

In practice, however, the performance of an alternative depends not only on the simulation logic but also on the input distributions used to represent stochastic primitives. These distributions are typically estimated from limited real-world data and are therefore subject to parameter estimation error or model misspecification, creating \emph{input uncertainty} (e.g., \citealt{song2014input, zhou2017simulation,  Xie2021nonparametric, lam2023statistical}). Such errors may propagate through the simulation model and even cause the alternative that appears best in simulation to be suboptimal in reality \citep{song2015input, zhou2015simulation}. 

Distributionally robust R\&S (DRR\&S), introduced by \citet{fan2013robust,fan2020distributionally}, addresses this issue by integrating R\&S with robust optimization and representing input uncertainty through a finite ambiguity set of plausible input distributions, each of which fits the available data well. Each distribution defines a scenario for every alternative, and the objective is to select the alternative with the best worst-case mean performance. Thus, with \(k\) alternatives and \(m\) plausible input distributions, DRR\&S involves \(km\) scenarios and has a two-layer minimax structure: the inner layer concerns the worst-case scenario within each alternative, while the outer layer compares alternatives through their worst-case means. This additional complexity has motivated a growing literature on DRR\&S procedures \citep[e.g.,][]{fan2013robust,zhang2016sequential,gao2017robust,shi2019worst,fan2020distributionally,wan2023upper,wan2024new}.
Other frameworks for R\&S under input uncertainty can be found in \citet{wang2025optimal}, \cite{song2025optimizing}, and the references therein.
Section~\ref{sec: problem} discusses how finite ambiguity sets may be constructed.

A central structural insight in  DRR\&S is \emph{additivity}. At first glance, the multiplicative \(km\)-scenario structure suggests that a good procedure needs to explore all scenarios sufficiently: it should learn the worst-case scenario within every alternative, while across alternatives, it determines which has the best worst-case performance. Interestingly, \citet{fan2013robust,fan2020distributionally} show that, when the sample size of each scenario is pre-specified (i.e., static), PICS admits an additive upper bound involving only \(k+m-2\) pairwise comparisons among \(k+m-1\) ``critical'' scenarios: all \(m\) scenarios of the best alternative and the worst-case scenario of each non-best alternative. 
Assuming that all scenario parameters are known, \citet{gao2017robust} derive oracle budget allocation ratios among scenarios by minimizing a multiplicative PICS bound as a surrogate objective; the resulting allocation also features a similar additive structure. \citet{wan2024new} further emphasize this structure by using the bound of \citet{fan2020distributionally} as an allocation objective.
If these critical scenarios were known, one could therefore concentrate most of the simulation effort on roughly \(k+m\), rather than \(km\), scenarios, to improve sample efficiency.

The challenge is that the critical scenarios are themselves unknown. Practical sequential procedures must learn them from simulation output while simultaneously allocating the sampling budget. A common approach is to repeatedly estimate the relevant performance parameters and myopically update the allocation according to the current sample information \citep[e.g.,][]{gao2017robust,wan2024new}, following a design principle that works well in traditional R\&S \citep[e.g.,][]{glynn2004large,chen2015ranking,gao2017new}. The static bounds and oracle allocations discussed above provide valuable allocation targets, but they do not characterize the sampling behavior of such implementable procedures. In particular, it remains unclear whether and how an adaptive, parameter-oblivious procedure can concentrate its budget on only \(k+m-1\) scenarios while maintaining the sufficient exploration needed for fundamental fixed-budget guarantees. This issue is intriguing because conventional R\&S requires every competing alternative to receive infinitely many observations. It is therefore not immediate and interesting to explore whether, and in what form, the additive concentration is compatible with the exploration required in DRR\&S.

There is a second, more subtle issue: which \(k+m-1\) scenarios should receive sustained sampling? Static and oracle analyses identify all scenarios of the true best alternative and the true worst-case scenario of every non-best alternative as critical. A natural interpretation is therefore that a sequential procedure must solve \(k\) implicit within-alternative worst-case scenario identification problems while resolving the outer-layer comparison. Yet our numerical investigation reveals more varied sample-path behavior. Although \citet{wan2024new} illustrate a sample path in which sampling concentrates on the hypothesized critical scenarios, we observe other sample paths in which some scenarios receiving sustained sampling lie outside this set, with their identities varying across sample paths; similar behavior occurs for the procedure of \citet{gao2017robust} (see Section~\ref{sec:existingsamplepath}). Rather than diminishing the role of the critical scenarios in oracle analysis, this behavior suggests that static criticality need not coincide with criticality along every sequential sample path. It therefore motivates examining whether identifying the true worst-case scenario of every non-best alternative, or accurately estimating its mean, is necessary for correct selection.

These observations lead to two questions that motivate this paper. First, \emph{how far can sequential sampling be concentrated while retaining fundamental performance guarantees} such as consistency and exponential PICS decay (Question 1)? Second, \emph{which scenarios must receive sustained sampling, and in particular, must they coincide with the conventionally defined critical scenarios} (Question 2)? We answer these questions by developing a notion of \emph{sequential additivity}, which characterizes additivity through the long-run sampling behavior induced by practical sequential procedures.


\emph{There exists a sharp sampling lower bound.}
We first establish an algorithm-independent necessary sampling condition: any nonanticipating sequential DRR\&S procedure that is consistent and has no access to the true parameters must sample at least $k+m-1$ scenarios infinitely often almost surely, including all $m$ scenarios of the best and at least one scenario of each non-best alternative. This result reveals a fundamental asymmetry between best and non-best alternatives and provides a theoretical benchmark for additivity. We then show that this lower bound is attainable by myopic sequential designs.

\emph{PICS convergence guarantees can be achieved.}
We abstract the common myopic design principle into a simplified, greedy additive allocation (AA) procedure to explicate the key structure and intuition. The procedure has two equal-allocation steps in each round: an $m$-step and a $k$-step. We first demonstrate the complementary roles of the two steps in exploration. Using boundary-crossing arguments tailored to their coupled adaptive sampling dynamics, we then derive a finite-budget upper bound on its PICS and show that this probability decays exponentially as the budget grows, thereby establishing consistency despite the procedure's greedy nature.

\emph{Additivity can be achieved in the strongest possible sense.}
With PICS convergence established, we proceed to characterize AA's sampling behavior. We show that, as the sampling budget tends to infinity, the AA procedure allocates infinitely many observations to exactly $k+m-1$ scenarios almost surely, while every other scenario receives only finitely many. AA therefore attains the necessary sampling lower bound established above, achieving additivity in the strongest possible sense. This result refines the usual R\&S sufficient-exploration intuition for DRR\&S.

\emph{Correct identification of worst-case scenarios may not be necessary.}
Moreover, we prove that when a non-best alternative has a non-worst-case scenario whose mean exceeds the worst-case mean of the true best, its true worst-case scenario has a positive probability of receiving only finitely many observations. Thus, the scenarios receiving sustained sampling need not coincide with the static critical set. This answers Question 2 and offers an intuition: if the true worst-case scenario appears unusually favorable early in the sampling process, another sufficiently unfavorable scenario may still reveal that the alternative is inferior. Therefore, identifying the true worst-case scenario, or estimating its mean, is not always necessary for non-best alternatives.

\emph{These insights extend to a broad class of procedures.}
Building on the additive structure revealed by AA, we introduce the general additive allocation (GAA) framework. It treats the \(m\)-step and $k$-step as modular R\&S subproblems and allows flexible sampling rules from traditional R\&S. Under suitable exploration conditions, we show that GAA procedures retain the key properties of AA. The framework therefore translates the structural analysis into a flexible approach for procedure design. 

\emph{Numerical experiments demonstrate the practical value of the GAA framework.}
In our experiments, GAA procedures generally outperform non-additive benchmarks, and new variants enabled by GAA can outperform existing ones that can be understood as GAA instances. We also investigate the use of common random numbers and find that their effect depends on the dimension along which dependence is introduced. 

Taken together, our results complement existing static and oracle perspectives by revealing how additivity emerges from the sample-path allocation behavior of sequential procedures. They clarify the different roles of best and non-best alternatives and of worst-case and non-worst-case scenarios, and translate these insights into a basis for designing new additive DRR\&S procedures. We also relate our findings to the traditional notion of rate-optimal budget allocation and outline directions for future work.

The remainder of this paper is organized as follows. Section~\ref{sec: problem} formulates the fixed-budget DRR\&S problem. Section~\ref{sec: analysis} introduces the AA procedure and analyzes its performance and budget allocation behavior, and Section~\ref{sec:AA property} establishes its additive properties. Section~\ref{sec: general} develops the GAA framework and studies its properties. Section~\ref{sec: numerical} presents the numerical experiments, Section~\ref{sec: conclusion} concludes the paper, and the E-Companion contains auxiliary technical details.

\section{Problem Formulation and Preliminaries}
\label{sec: problem}

Suppose that there are \(k\geq2\) alternatives, denoted by \(\mathcal K=\{1,\ldots,k\}\), and that the objective is to identify the alternative with the smallest mean performance through simulation. Suppose also that the input data admit \(m>1\) plausible input models \(P_j\), \(j=1,\ldots,m\), which form a finite ambiguity set \(\mathcal P\). Following \citet{fan2020distributionally}, such an ambiguity set may be constructed by fitting candidate distribution families to the available data and retaining those that are not rejected by a goodness-of-fit test.  Each input model, or input distribution, specifies the distributions governing all stochastic primitives of the simulation and may therefore comprise several component distributions. Large bootstrap-generated collections of closely related distributions, for which similarity may be informative, are beyond the scope of this paper \citep{Xie2021nonparametric}. 

We use a scenario \((i,j)\) to represent alternative \(i\) under input distribution \(P_j\).  For each scenario $(i,j)$, where $i=1, \dots, k$ and $ j=1, \dots, m$, we use the random variable $X_{ij}$ to denote the simulated output, with true mean \(\mu_{ij} = \E[X_{ij}]\) and positive variance \(\sigma_{ij}^2 = \mathrm{Var}[X_{ij}] > 0\).  Following the convention in the R\&S literature, we assume that all $X_{ij}$ are normally distributed, i.e., $X_{ij} \sim \mathcal{N}(\mu_{ij}, \sigma_{ij}^2)$, and that all simulation observations are mutually independent both within and across scenarios. We explore the effect of CRNs and discuss non-normal distributions in Section \ref{sec: numerical}.
Given the ambiguity set, DRR\&S takes an ambiguity-aversion perspective, aiming to select the \emph{minimax} best alternative with the smallest worst-case mean performance across the input distributions, i.e., alternative 
$
i^* = \argmin_{i \in \mathcal{K}} \max_{P_j \in \mathcal{P}} \ \mu_{ij}.
$

For notational convenience, we hypothetically relabel the scenarios within each alternative so that scenario \(1\) is its true worst-case scenario, namely, \(\mu_{i1}=\max_{j=1,\cdots, m}\mu_{ij}\).  We allow tied scenario means within each alternative. These local labels need not correspond to the same ordering of input distributions across alternatives.
We label alternative \(1\) as the unique best alternative and alternative \(2\) as a second-best alternative, so that \(\mu_{11}<\mu_{21}=\min_{i\geq2}\mu_{i1}\). These labels are used only to simplify notations and are neither known nor required by the procedures.
We use $\delta=\mu_{21} - \mu_{11}>0$ to denote the unknown robust optimality gap.

We adopt a fixed-budget formulation in which the total sampling budget \(N\) is predetermined. A DRR\&S procedure allocates this budget among the scenarios, and after exhausting it, selects an alternative \(\hat{b}\) based on the collected sampling information. The performance of a procedure is measured by the probability of correct selection (PCS), defined as $\Pr \left\{\hat{b} = 1 \right\}$.
A fundamental property of any fixed-budget R\&S procedure is consistency \citep{hong2021review}, also known as asymptotic optimality \citep{frazier2008knowledge}. Intuitively speaking, it assures that a correct selection can be made with probability one when the total simulation effort is large enough. 
We formalize it as follows:
 
\begin{definition}
\label{def:consistency}
\itshape A DRR\&S procedure is consistent over the problem class described above if, for every problem instance in the class, its PCS satisfies $\lim_{N \rightarrow \infty} \mathrm{PCS} = 1$.
\end{definition}
 
Achieving consistency alone is not an inherently challenging task. A simple sufficient condition is to sample every scenario infinitely often as \(N\) increases to infinity. 
A naïve equal allocation procedure, for example, satisfies this condition.
  In this paper, we take a step further. We rigorously investigate issues around additivity in allocating the total sampling budget. We begin by establishing an algorithm-independent necessary sampling lower bound. Temporarily, let \(n_{ij}(N)\) denote the number of observations allocated to scenario $(i,j)$ when the total budget \(N\) is exhausted. 
 
\begin{theorem}[Necessary $k+m-1$ Sampling Condition]
    \label{thm:necessary-sampling}
    Consider any nonanticipating DRR\&S procedure that is consistent and has no access to the true parameters. Fix a problem instance in which the second-best alternative is unique, so that, when $k > 2$, 
    \(\mu_{21}<\min_{i\geq3}\mu_{i1}\). Then, almost surely,
    \[
    \lim_{N\to\infty}n_{1j}(N)=\infty,
    \quad j=1,\ldots,m,
    \quad\text{and}\quad
    \lim_{N\to\infty}\sum_{j=1}^m n_{ij}(N)=\infty,
    \quad i=2,\ldots,k.
    \]
    \end{theorem}


Theorem~\ref{thm:necessary-sampling} shows that consistency requires all scenarios of the best alternative and at least one scenario of every non-best alternative to be sampled infinitely often.  This result highlights a fundamental asymmetry between the best and non-best alternatives and provides a benchmark for how additive a consistent procedure can possibly be. Intuitively, 
changing a single scenario can destroy the optimality of the best alternative, so every one of its scenarios must be sufficiently sampled. By contrast, making a non-best alternative optimal may require changing all its scenarios.
The unique-second-best condition is technical and is used only in the proof’s change-of-measure argument. See Section~\ref{subsec:sampling_lower_bound} for details.

\section{A Simplified Procedure and Its Budget Allocation Analysis}
\label{sec: analysis}
We introduce the additive allocation (AA) procedure in Section \ref{subsec: additive_allocation} and illustrate the complementary roles of its two allocation steps in Section \ref{subsec: insights}. We then analyze its PCS and budget allocation behavior in Section \ref{subsec:boundary}, laying the foundation for establishing its additive properties in Section \ref{sec:AA property}.
 
\subsection{Myopic and Sequential: The Additive Allocation Procedure}
\label{subsec: additive_allocation}
Prior discussions on additivity identify \(k+m-1\) ``critical'' scenarios: all \(m\) scenarios of the best alternative and the worst-case scenario of each non-best alternative. Guided by this intuition, a common sequential heuristic is to conduct budget allocation on the current ``critical'' scenarios (identified based on the current sample information) in each round \citep[e.g.,][]{gao2017robust, wan2024new} — a prevalent myopic practice that works well in R\&S \citep[e.g.,][]{glynn2004large, chen2015ranking, gao2017new}. We isolate this principle in a simplified procedure, referred to as the AA procedure, to explicate the key structure.
 
AA begins by drawing one observation from every scenario to initialize the sample means. In each subsequent round, it identifies the empirical worst-case scenario of every alternative and the current empirical best alternative. It then performs two steps. The \(m\)-step samples every scenario of the current best alternative once, while the \(k\)-step samples the empirical worst-case scenario of each of the other \(k-1\) alternatives once. Thus, exactly \(k+m-1\) current ``critical'' scenarios are sampled in each round.

After the total budget is exhausted, we intuitively let AA select the alternative that has been identified as the current best most frequently, i.e., the one that has received the most \(m\)-steps. Because \(m>1\), this is also the alternative with the largest sample size, echoing early simulation optimization work \citep[see, e.g.,][]{andradottir2009balanced}. Sample-mean-based selection rules could also be considered, but this rule is sufficient for our purpose. Procedure \ref{algo:additive_rrs} provides the details with ties broken arbitrarily.

\vspace{-10pt}
\begin{algorithm}[htbp]
\SetAlgoCaptionLayout{centerline}
\caption{\textbf{Additive Allocation (AA)}}
\label{algo:additive_rrs}
\KwIn{$k$ alternatives, $m$ scenarios per alternative, total sampling budget $N$.}
For each scenario $(i,j)$, take one observation $x_{ij}$, let sample size $n_{ij} = 1$ and sample mean $\bar{X}_{ij}(1)=x_{ij}$\;
\For{$t = 1$ \KwTo $\left\lfloor(N - mk)/(m + k - 1)\right\rfloor$}{
    Identify the empirical worst-case scenario $\hat{1}_i = \argmax_{j=1, \dots, m} \bar{X}_{ij}(n_{ij})$ for each alternative $i $\;
    Identify the current best alternative $\hat{b} \gets \argmin_{i=1, \dots, k} \bar{X}_{i\hat{1}_i}(n_{i\hat{1}_i})$\;
    \textit{\textbf{$m$-step:}} for each scenario $(\hat{b},j)$ of the alternative $\hat{b}$, take one observation $x_{\hat{b}j}$, update its sample mean $\bar{X}_{\hat{b}j}(n_{\hat{b}j}+1)=(n_{\hat{b}j} \bar{X}_{\hat{b}j}(n_{\hat{b}j}) + x_{\hat{b}j}) / (n_{\hat{b}j} + 1)$, and set $n_{\hat{b}j} \gets n_{\hat{b}j} + 1$\;
    \textit{\textbf{$k$-step:}} for each alternative $i \in \{1, \dots, k\} \setminus \{\hat{b}\}$, take one observation $x_{i\hat{1}_i}$ from scenario $(i, \hat{1}_i)$, update its sample mean $\bar{X}_{i \hat{1}_i}(n_{i \hat{1}_i}+1)=(n_{i \hat{1}_i} \bar{X}_{i\hat{1}_i}(n_{i\hat{1}_i}) + x_{i\hat{1}_i}) / (n_{i \hat{1}_i} + 1)$, and set $n_{i \hat{1}_i} \gets n_{i \hat{1}_i} + 1$\;
}
\vspace{6pt}
\KwOut{The alternative that is most frequently chosen for the $m$-steps.}
\end{algorithm}
\vspace{-10pt}

We start with this procedure because of its structural simplicity. When $m=1$, DRR\&S reduces to the traditional R\&S problem, and AA naturally degenerates into the simplest equal allocation procedure. This simplicity may provide convenience for investigating structural properties, through which we wish to answer the central questions raised in Section \ref{sec: intro}. Although AA is deliberately simplified for tractable analysis, our later results show that its insights generalize to a broad class of procedures. Notice that despite AA's simple allocation rule, its adaptive sample path makes its performance and budget allocation behavior nontrivial to analyze. We address this using a sample-path argument based on boundary crossings.

\subsection{The Complementary Roles of the $m$-Step and $k$-Step}
\label{subsec: insights}

\begin{figure}[htp]
     \FIGURE
    {\includegraphics[width=0.6\linewidth]{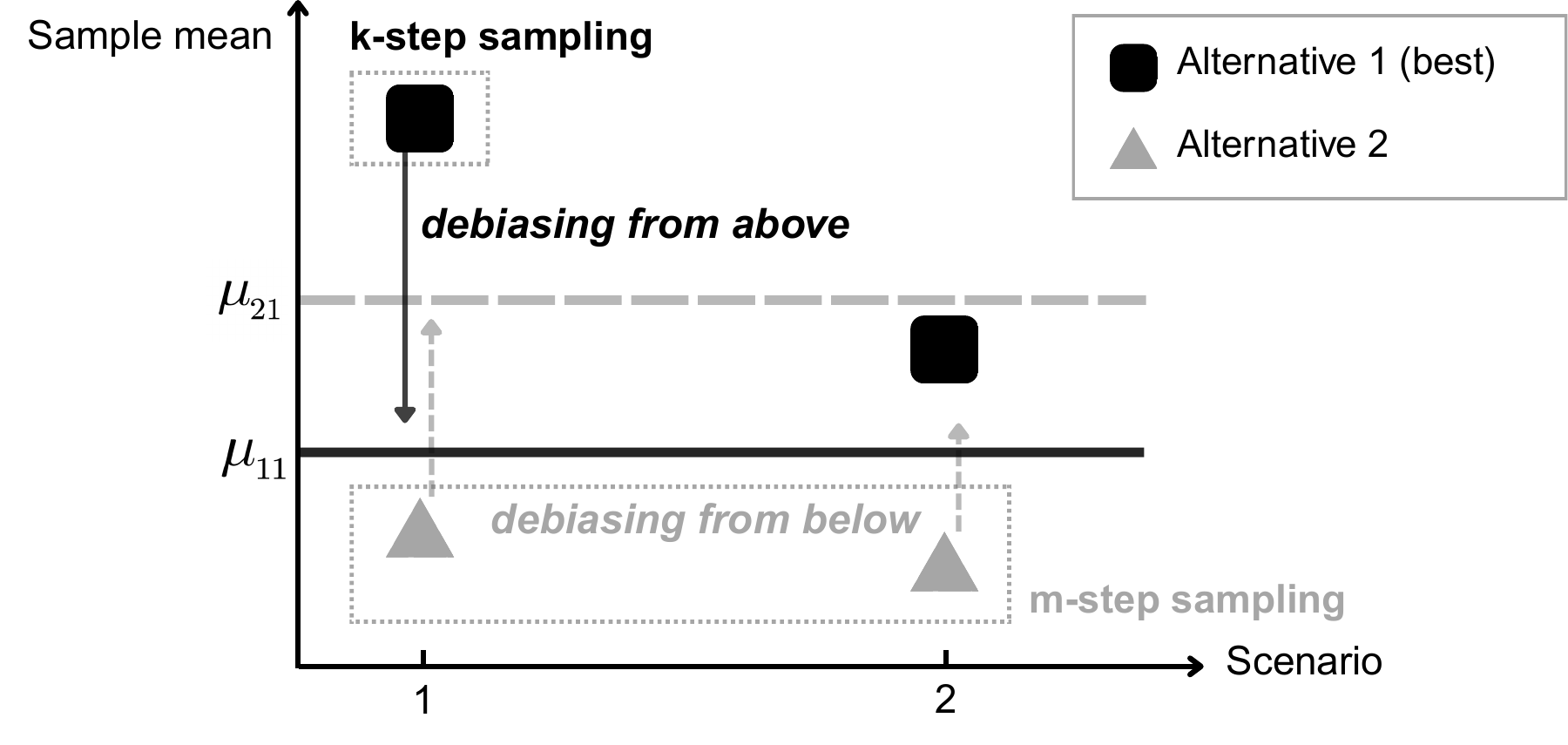}}
     {Roles of the $m$-step and $k$-step in a Statistically Adverse Round of a Two-Alternative Problem \vspace{10pt}    \label{fig:illustration}}
    {This example illustrates a statistically adverse round where the worst-case sample mean of alternative 1 is overly pessimistic (too high), while that of alternative 2 is overly optimistic (too low). As a result, alternative 2 is selected for the $m$-step.}
\end{figure}

Before proceeding to the analysis, we illustrate the roles of the $m$-step and $k$-step in the AA procedure. Given a robust-\emph{minimum}-seeking DRR\&S problem, a statistically ideal round is one in which the worst-case sample mean of the best alternative is lower than those of all other alternatives, so that the current empirical best truly corresponds to the true best. However, without such luck, this may not hold. In procedure design, we must account for the adverse scenario where the worst-case sample mean of the best is pessimistic (too high), while that of a non-best alternative is optimistic (too low). In this case, the current best may not be the true best. The AA procedure addresses this issue through the complementary actions of the $m$-step and $k$-step. For the optimistic non-best, it will be sampled in an $m$-step, pushing its sample means up, i.e., debiasing from below. The $k$-step explores the current worst-case scenario of the true best alternative, which appears pessimistic, in order to reduce its sample mean — i.e., debiasing from above. See Figure~\ref{fig:illustration} for a visual illustration of these roles. From this viewpoint, we can see that both steps are necessary.  

\subsection{Bounding the Rounds: A Boundary-Crossing Perspective}
\label{subsec:boundary}

In this subsection, we analyze the AA procedure with the goal of characterizing the number of rounds it requires to ensure a correct selection. This analysis will yield a lower bound on the PCS under a finite sampling budget and provide structural insights into how the budget is allocated among the alternatives. To achieve this, we adopt a sample-path analysis based on boundary-crossing times for both the best and non-best alternatives. We first introduce the concept of last exit times, which record the last sample sizes at which a scenario's sample mean lies above or below a specified threshold.

\subsubsection{Last Exit Times.}
\label{subsubsec:last_exit_times}
For each scenario $(1,j)$ of the best alternative (alternative 1) with mean $\mu_{1j}$, we define the last exit time from a boundary $b > \mu_{1j}$ as
$$
U_{1j}(b) = \sup \{ n \geq 1 : \bar{X}_{1j}(n) \geq b \},
$$
where $\bar{X}_{1j}(n)$ is the sample mean of scenario $(1,j)$ after $n$ observations. By convention, if the sample mean process $\bar{X}_{1j}(n)$ stays strictly below $b$ for all $n$, we set $U_{1j}(b) := 0$. Similarly, for each scenario $(i,j)$ of a non-best alternative $i \in \{2, \dots, k\}$ and any boundary $b$, we define
$$
L_{ij}(b) = \sup \{ n \geq 1 : \bar{X}_{ij}(n) \leq b \},
$$
where $L_{ij}(b) := 0$ if the sample mean process $\bar{X}_{ij}(n)$ remains strictly above $b$ for all $n$. This last exit time may be infinite; it is finite almost surely when $b<\mu_{ij}$.

Intuitively, $U_{1j}(b)$ records the last sample size at which the sample mean of scenario $(1,j)$ is at or above $b$. When $b<\mu_{ij}$, $L_{ij}(b)$ similarly records the last sample size at which the sample mean of scenario $(i,j)$ is at or below $b$. Importantly, these last exit times are defined with respect to the entire sample mean process $\{\bar{X}_{ij}(n)\}_{n=1}^\infty$ and are independent of the total sampling budget. Their definition does \emph{not} require the scenario to be sampled infinitely often by the procedure. As we shall see, they provide a clean sample-path bound sufficient for a correct final selection.
\subsubsection{Counters.} We next define counters to describe the sampling process of AA.
Given a total sampling budget $N \geq mk$, let $T=\left\lfloor(N - mk)/(m + k - 1)\right\rfloor$ denote the number of completed rounds, and let $t=1,2,\dots,T$ index these rounds. Each round consists of one $m$-step and one $k$-step. Since each scenario $(i, j)$ is sampled once during the initialization stage, we let $n_{ij}(0) = 1$ for all $i = 1, \dots, k$ and $j = 1, \dots, m$. For each $t$, define $I_i^{m}(t) = 1$ if alternative $i$ is selected for the $m$-step in round $t$, and $0$ otherwise; $I_i^{k}(t) = 1$ if alternative $i$ is selected for the $k$-step in round $t$, and $0$ otherwise. For each alternative $i$, we define $n_i^m(t) = \sum_{\tau = 1}^{t} I_i^{m}(\tau)$ and $n_i^{k}(t) = \sum_{\tau = 1}^{t} I_i^{k}(\tau)$ as the number of $m$-steps and $k$-steps, respectively, in which alternative $i$ is sampled up to and including round $t$; these are referred to as the allocated $m$-steps and $k$-steps for alternative $i$. 
We also define $n_{ij}(t)$ as the total number of observations allocated to scenario $(i, j)$ after $t$ rounds.

\subsubsection{Boundary-Crossing Analysis.} Now, we examine how the AA procedure behaves through the lens of the last exit times. We consider a boundary $b_{\delta} \in (\mu_{11}, \mu_{21})$. Since $\mu_{21} - \mu_{11} = \delta > 0$, such a boundary always exists. 
For every non-best alternative $i$, $\mu_{i1}\geq\mu_{21}>b_\delta$, so $L_{i1}(b_\delta)$, and hence $\min_{j=1,\dots,m}L_{ij}(b_\delta)$, is finite almost surely.
This leads to the following observations. Observation \ref{obs: non-best} characterizes the number of $m$-steps allocated to each non-best alternative, while Observations \ref{obs: best-below} and \ref{obs: best-above} characterize the number of $k$-steps allocated to the best alternative. Together, these observations will lead to Lemma \ref{lem: sufficient_time} and the number of rounds sufficient for the AA procedure to ensure a correct selection. 

\begin{observation}
\label{obs: non-best}
At round $t$, for each non-best alternative $i$, if the number of $m$-steps allocated to $i$ satisfies
$$
n_i^m(t)+1 > \min_{j = 1, \dots, m} L_{ij}(b_{\delta}),
$$
then its empirical worst-case (maximum) sample mean across scenarios will remain permanently above $b_{\delta}$ for all subsequent rounds, i.e., 
\[
\max_{j = 1, \dots, m} \bar{X}_{ij}(n_{ij}(t^\prime)) > b_\delta, \, \forall t^\prime \geq t.
\]
The ``+1'' accounts for the initial one observation.
\end{observation}

For the best alternative (alternative $1$), to analyze its allocated number of $k$-steps, we divide the rounds $t$ into two groups: (1) rounds where, at the beginning of the round, its maximum sample mean is below the boundary $b_\delta$, and (2) rounds where, at the beginning of the round, its maximum sample mean is above $b_\delta$. Due to the normality assumption, it is a probability-zero event that the worst-case sample mean hits $b_\delta$.

First, let's consider group (1). When the maximum sample mean of the best alternative is below $b_\delta$, it can only be selected for a $k$-step if a certain non-best alternative also has its maximum sample mean below $b_\delta$ and is selected for an $m$-step.
By Observation \ref{obs: non-best}, in this case, for each non-best alternative $i = 2, \dots, k$, the total number of such $m$-steps is upper bounded by $\min_{j=1, \dots, m} L_{ij}(b_\delta)$. This leads to the following observation:
\begin{observation}
\label{obs: best-below}
Regardless of the total number of rounds $T=\left\lfloor(N - mk)/(m + k - 1)\right\rfloor$, the total number of $k$-steps allocated to alternative 1 while its worst-case sample mean is below the boundary $b_\delta$ satisfies
$$
\sum_{\tau=1}^{T} I_1^{k}(\tau) \, \mathbbm{1}_{ \max_{j=1,\dots, m} \bar{X}_{1j}(n_{1j}(\tau-1)) < b_\delta}
\le \sum_{i=2}^k \min_{j=1, \dots, m} L_{ij}(b_\delta).
$$
Let $\mathbbm{1}_{A}$ denote the indicator function, which equals 1 if condition $A$ holds and 0 otherwise. 
\end{observation}

Next, let's turn to group (2). For each scenario $(1,j)$ of the best alternative, note that after the initial one and $U_{1j}(b_\delta)$ additional observations, the sample mean will remain permanently below the threshold $b_\delta$. This leads to the following observation:
\begin{observation}
\label{obs: best-above}
For the best alternative (alternative 1), regardless of the total number of rounds $T=\left\lfloor(N - mk)/(m + k - 1)\right\rfloor$,  the total number of $k$-steps allocated while its worst-case sample mean is above the boundary $b_\delta$ satisfies
$$
\sum_{\tau=1}^{T} I_1^{k}(\tau) \, \mathbbm{1}_{\max_{j=1,\dots, m}  \bar{X}_{1j}(n_{1j}(\tau-1)) >  b_\delta}
\le  \sum_{j=1}^m U_{1j}(b_\delta).
$$
\end{observation}


From Observations \ref{obs: best-below} and \ref{obs: best-above}, and noting that each round of the AA procedure includes one $m$-step and one $k$-step, we obtain the following result.
\begin{lemma}[Bounds on the Steps]
\label{lem: sufficient_time} 
For every $T$, the following hold almost surely for the AA procedure:
\begin{itemize}
    \item [(1)] for alternative 1, its allocated number of $k$-steps satisfies
\begin{eqnarray}
    n_1^k(T) \leq S(b_\delta) := \sum_{i=2}^k \min_{j=1, \dots, m} L_{ij}(b_\delta) + \sum_{j=1}^m U_{1j}(b_\delta);
\end{eqnarray}
\item [(2)] the total number of $m$-steps allocated to the non-best alternatives satisfies 
\begin{eqnarray}
     \sum_{i=2}^k n_i^m(T) = n_1^k(T) \leq S(b_\delta).
\end{eqnarray}
\end{itemize}
\end{lemma}

Lemma \ref{lem: sufficient_time} immediately provides an upper bound on the number of rounds required for the AA procedure to achieve a correct selection. Recall that, after completing all full rounds allowed by the total sampling budget, the alternative with the most $m$-steps is selected as the best. Therefore, on any sample path for which $T > 2S(b_\delta)$, AA selects the true best alternative. This is because, when $T > 2S(b_\delta)$,
$$
n_1^m(T) = T - \sum_{i=2}^k n_i^m(T) > S(b_\delta) \geq \sum_{i=2}^k n_i^m(T) \geq n_i^m(T), \, \forall i \geq 2.
$$
Consequently, we obtain the following result on the PCS of the AA procedure.
\begin{lemma}[A PCS Lower Bound]
\label{lem: PCS-bound}
Given a total sampling budget $N \geq mk$ and for any $b_\delta \in (\mu_{11}, \mu_{21})$, the PCS of the AA procedure satisfies
\begin{eqnarray*}
        \mathrm{PCS} \geq 
    \Pr \left\{ \left\lfloor\frac{N - mk}{m + k - 1}\right\rfloor > 2 
    \left( \sum_{i=2}^k \min_{j=1, \dots, m} L_{ij}(b_\delta) 
    + \sum_{j=1}^m U_{1j}(b_\delta) \right) \right\}.
\end{eqnarray*}
\end{lemma}
\vspace{4pt}

Lemmas \ref{lem: sufficient_time} and \ref{lem: PCS-bound} provide a clear characterization of the budget allocation behavior and the PCS of the AA procedure under a finite total sampling budget. These results serve as the foundation for our subsequent analysis of the properties and additive structure of the AA procedure. To the best of our knowledge, this is the first such analysis conducted for DRR\&S procedures.

\section{Properties of the Additive Allocation Procedure}
\label{sec:AA property}
Building on the preceding boundary-crossing analysis, we now establish the performance guarantees and additive structure of the AA procedure. Section~\ref{subsec: properties_last_exit} develops the properties of last exit times needed for our analysis. Section~\ref{subsec: consistency} establishes an exponential decay of the PICS and consistency, while Section~\ref{subsec: additive} characterizes the asymptotic budget allocation and proves sequential additivity.

\subsection{Properties of Last Exit Times}
\label{subsec: properties_last_exit}
Lemmas~\ref{lem: sufficient_time} and~\ref{lem: PCS-bound} depend on the scenario-specific last exit times $U_{1j}(b)$ and $L_{ij}(b)$. To characterize their probabilistic behavior, we consider a generic standardized sample-mean process. Let $Z_1,Z_2,\dots$ be i.i.d. standard normal random variables, and define
$\bar Z(n)=n^{-1}\sum_{i=1}^n Z_i$.
For a boundary $b>0$, let
\begin{eqnarray}
\label{eq: general_def}
U(b)=\sup\{n\geq1:\bar Z(n)\geq b\}
\quad\text{and}\quad
L(-b)=\sup\{n\geq1:\bar Z(n)\leq-b\}.
\end{eqnarray}
By symmetry of the standard normal distribution, $U(b)$ and $L(-b)$ are identically distributed. After centering and scaling, $U_{1j}(b_\delta)$ corresponds to $U((b_\delta-\mu_{1j})/\sigma_{1j})$, while $L_{ij}(b_\delta)$ corresponds to $L(-(\mu_{ij}-b_\delta)/\sigma_{ij})$ whenever $\mu_{ij}>b_\delta$. The following lemma provides an exponential bound on their tail probabilities. Its proof is provided in Section~\ref{subsec: proof_lemma_lastexit_tail}.
\begin{lemma}[Tail Probabilities]
\label{lem: lastexit_tail}
For any $n\in\mathbbm N^+$ and $b>0$,
$
\Pr\{U(b)>n\}
=
\Pr\{L(-b)>n\}
\leq 2\exp(-nb^2/2).
$
\end{lemma}

The tail bound immediately implies that both last exit times are finite almost surely. We state this consequence separately because it will be used repeatedly below.
\begin{lemma}[Almost Sure Finiteness]
\label{lem: finite}
For $b>0$, the last exit times defined in Equation~\eqref{eq: general_def} satisfy
$
\Pr\{U(b)<\infty\}
=
\Pr\{L(-b)<\infty\}
=1.
$ 
\end{lemma}

\subsection{Convergence of PICS}
\label{subsec: consistency}
Combining Lemmas~\ref{lem: PCS-bound} and~\ref{lem: lastexit_tail}, we show that the PICS of the AA procedure decays at an exponential rate as $N$ increases, as formalized in the following proposition. The proof is provided in Section \ref{subsec: proof_exponential_decay}.

\begin{proposition}[Exponential Decay of PICS]
\label{prop: exponential_decay}
 For any threshold $b_\delta \in (\mu_{11}, \mu_{21})$, define $M_i = \left| \left\{ j : \mu_{ij} > b_\delta \right\} \right|$ as the number of scenarios of alternative $i$ whose true means exceed the boundary $b_\delta$. By definition, $M_i \geq 1$ for $i=2, \dots,k$. Let $q=m+k-1$, $T=\lfloor(N-mk)/q\rfloor$, and $r=\max\{0,\lceil T/(2q)\rceil-1\}$. Then, for any total sampling budget $N \geq mk$, the PICS of the AA procedure satisfies 
\begin{eqnarray*}
\mathrm{PICS} \le \sum_{i=2}^k \left[ 2^{M_i} \exp\left( -r \sum_{j: \mu_{ij} > b_\delta} \frac{(\mu_{ij}-b_\delta)^2}{2\sigma_{ij}^2} \right) \right] + \sum_{j=1}^m \left[ 2 \exp\left( -r \frac{(b_\delta - \mu_{1j})^2}{2\sigma_{1j}^2} \right) \right].
\end{eqnarray*}
\end{proposition}

 An exponential upper bound on the PICS, such as that in Proposition~\ref{prop: exponential_decay}, is desirable in R\&S because it guarantees substantial gains from increasing the sampling budget. Such guarantees, however, are often difficult to establish for adaptive procedures. Although our bound may be conservative, to the best of our knowledge, Proposition~\ref{prop: exponential_decay} provides the first exponential PICS bound for a sequential procedure in the DRR\&S setting. Furthermore, the bound involves all scenarios of the best alternative but, for each non-best alternative, only those scenarios whose true means exceed the chosen boundary \(b_\delta\). This dependence hints at a form of sparsity: only scenarios capable of separating a non-best alternative from the best alternative contribute to its guaranteed decay exponent.
    \citet{gao2017robust}, \citet{fan2020distributionally}, and \citet{wan2024new} derive tighter oracle PICS bounds for optimized static allocation ratios when parameters of scenarios are known. 
    The resulting allocations, however, are not directly implementable when those parameters must be learned through simulation. See Section~\ref{subsec: fixed_asymptotic_budget_allocation_ratios} for a discussion on how to achieve such ratios asymptotically.

Moreover, Proposition \ref{prop: exponential_decay} immediately implies that the AA procedure is consistent under the fixed-budget formulation, as summarized in the following theorem.
\begin{theorem}[Consistency]
\label{thm: consistency} The AA procedure is consistent, i.e., $\lim_{N \rightarrow \infty} \mathrm{PCS} = 1$.
\end{theorem}

Although consistency itself is not difficult to obtain, it is notable here because AA operates in a fundamentally greedy manner. The complementary roles of the \(m\)-step and \(k\)-step, as discussed in Section~\ref{subsec: insights}, provide sufficient exploration despite this greedy design.

\subsection{Additive Properties}
\label{subsec: additive}

Having established the performance guarantees of AA, we now characterize its additive structure as the total sampling budget $N\to\infty$. Recall from Lemma~\ref{lem: sufficient_time} that the total number of $m$-steps allocated to the non-best alternatives is bounded by a sum of last exit times, which is finite almost surely by Lemma~\ref{lem: finite}. The best alternative must therefore receive infinitely many $m$-steps, implying that each of its scenarios is sampled infinitely often. Meanwhile, every non-best alternative is sampled through either an $m$-step or a $k$-step in every round, so its overall sample size also diverges. The following proposition formalizes these observations. Its proof is provided in Section~\ref{subsec: proof_prop_allocation}.
 
\begin{proposition}[Asymptotic Budget Allocation]
\label{prop: allocation}
For the AA procedure, it holds almost surely that
    \begin{enumerate}
        \item For $j=1, \dots, m$, $\lim_{N\rightarrow \infty} n_{1j} = \infty$. 
        \item For alternative $i=2, \dots, k$, $\lim_{N\rightarrow \infty} \sum_{j=1}^m n_{ij} = \infty$.
    \end{enumerate}
    
\end{proposition}

A missing piece in the above analysis is the asymptotic budget allocation across the scenarios of each non-best alternative. A natural prior belief—given the consistency of the procedure—is that every scenario of a non-best alternative should eventually receive an infinite number of observations. However, the following theorem shows that this is not necessary. Its proof is provided in Section~\ref{subsec: proof_thm_additive}.

\begin{theorem}[Sequential Additivity]
\label{thm: additive}
For the AA procedure, it holds almost surely that
\[
\sum_{i=1}^k\sum_{j=1}^m
\mathbbm{1}_{\lim_{N\rightarrow\infty}n_{ij}(N)=\infty}
=k+m-1.
\]

\end{theorem}

Combined with Proposition~\ref{prop: allocation}, Theorem~\ref{thm: additive} implies that all $m$ scenarios of the best alternative and exactly one scenario of each non-best alternative receive infinitely many observations. AA therefore attains the necessary $k+m-1$ sampling lower bound in Theorem~\ref{thm:necessary-sampling} and is as additive as any consistent procedure can possibly be. Asymptotically, it focuses exclusively on just $k + m - 1$ ``critical'' scenarios out of the total $km$, while still achieving the above performance guarantees. 

The intuition behind this result stems from Lemmas~\ref{lem: sufficient_time} and~\ref{lem: finite}, which together imply that the non-best alternatives receive only finitely many $m$-step allocations. Beyond that point, each of them is sampled only via $k$-steps, which operate greedily—always sampling the current worst-case scenario with the maximal sample mean among scenarios. Within each alternative, this greedy mechanism does not allow infinite switching between scenarios: only one scenario becomes the current worst-case scenario infinitely often and continues to be selected and sampled. All other scenarios are effectively ignored in the limit.


This raises a sharper question: which scenario of each non-best alternative is the ``critical'' one that ultimately receives an infinite number of observations? Based on the conventional intuition behind additivity, one might expect it to be the true worst-case scenario, which would require correctly identifying that scenario during the sampling process. The following results show otherwise.
For any boundary $b_\delta$, let $a_{ij}=\Phi\left((b_\delta-\mu_{ij})/\sigma_{ij}\right)$ and $b_{ij}=1-\exp\left(-(b_\delta-\mu_{ij})^2/(2\sigma_{ij}^2)\right)$ for every scenario $(i,j)$. Notice that $a_{ij} >0$ and  $b_{ij} >0$ when $b_\delta \neq \mu_{ij}$. The proofs are provided in Section~\ref{subsec: proof_thm_non_necessity}.

\begin{proposition}[Possible Limiting Scenarios]
    \label{prop: possible_limiting}
     For the AA procedure, fix any non-best alternative $i=2,\dots,k$ and any scenario $j$ such that $\mu_{ij}>\mu_{11}$. For any $b_\delta\in(\mu_{11},\mu_{ij})$,
    \begin{eqnarray*}
         &&\Pr\left\{\lim_{N\rightarrow\infty}n_{ij}(N)=\infty,\ 
         \lim_{N\rightarrow\infty}n_{i\ell}(N)<\infty,\ \forall\,\ell\neq j\right\}  \geq b_{ij}\prod_{\ell\neq j}a_{i\ell}\prod_{\ell=1}^{m}b_{1\ell}>0.
    \end{eqnarray*}
\end{proposition}

\begin{theorem}[Non-Necessity of Correct Identification of Worst-Case Scenarios]
\label{thm: non_necessity}
For a non-best alternative $i$, if there exists a non-worst-case scenario $j$ such that $\mu_{11}<\mu_{ij}<\mu_{i1}$, then, for any $b_\delta\in(\mu_{11},\mu_{ij})$,
\begin{eqnarray*}
\Pr\left\{\lim_{N\rightarrow\infty}n_{ij}(N)=\infty,\ 
\lim_{N\rightarrow\infty}n_{i\ell}(N)<\infty,\ 
\forall\,\ell\text{ such that }\mu_{i\ell}=\mu_{i1}\right\}
\geq b_{ij}\prod_{\ell\neq j}a_{i\ell}\prod_{\ell=1}^{m}b_{1\ell}>0.
\end{eqnarray*}
\end{theorem}

Proposition~\ref{prop: possible_limiting} shows that every scenario of a non-best alternative whose mean exceeds the worst-case mean of the best alternative has a positive probability of being the unique scenario sampled infinitely often. Theorem~\ref{thm: non_necessity} strengthens this implication: when such a scenario is not a true worst-case scenario, there is a positive probability that no true worst-case scenario of that alternative is sampled infinitely often. This result departs from the conventional view in the DRR\&S literature that identifying the worst-case scenario \citep{fan2013robust, fan2020distributionally, gao2017robust} or accurately estimating the worst-case mean \citep{wan2023upper} is essential.
For intuition, the empirical worst-case mean of the best alternative acts as a ``guide rail.'' A non-best alternative need only have one scenario whose sample mean remains above this guide rail to prevent that alternative from appearing best. This scenario need not be a true worst-case scenario. Therefore, correctly identifying or tracking the true worst-case scenarios of a non-best alternative may be unnecessary.

\section{Generalization}
\label{sec: general}
In this section, we explore extensions to the AA procedure that aim to generalize the insights and enhance practical performance while preserving its additive structure. Specifically, in Section~\ref{subsec: general_additive}, we present a generalized framework for additive allocation procedures, which allows for leveraging existing fixed-budget R\&S procedures in a modular fashion. Then, in Section~\ref{subsec: general_properties}, we discuss properties of this generalized class.

\subsection{General Procedure Design}
\label{subsec: general_additive}
The AA procedure is built around the $m$-step and $k$-step performed in each round. The boundary-crossing analysis in Section~\ref{subsec:boundary} reveals how their complementary roles produce the additive structure. AA uses equal allocation within both steps for clarity, but this simple rule can be conservative in practice. We therefore retain the myopic, round-by-round structure of AA while allowing adaptive allocation within each step. This leads to the general additive allocation (GAA) procedure in Procedure~\ref{algo:general_additive_rrs}.

\begin{algorithm}[htbp]
\SetAlgoCaptionLayout{centerline}
\caption{\textbf{General Additive Allocation (GAA) }}
\label{algo:general_additive_rrs}
\KwIn{$k$ alternatives, $m$ scenarios per alternative, total sampling budget $N$, initial sample size $n_0$, $m$-step sampling rule $\mathcal{M}$, $\Delta_m \geq 1$, $k$-step sampling rule $\mathcal{K}$, and $\Delta_k \geq 1$.}
For each scenario $(i,j)$, take $n_0$ observations to initialize the sample mean $\bar{X}_{ij}(n_0)$ and sample standard deviation $\hat \sigma_{ij}(n_0)$; set $n_{ij} \gets n_0$, $n_{ij}^m \gets 0$, $n_{ij}^k \gets 0$, $r_i^m \gets 0$, and $r_i^k \gets 0$\;
\While{$\sum_{i=1}^{k} \sum_{j=1}^{m} n_{ij}+ \Delta_m + \Delta_k \leq N$}{
    Identify the current worst scenario $\hat{1}_i = \argmax_{j=1, \dots, m} \bar{X}_{ij}(n_{ij})$ for each alternative $i = 1, 2, \dots, k$\; 
    Identify the current best alternative $\hat{b} \gets \argmin_{i=1, \dots, k} \bar{X}_{i\hat{1}_i}(n_{i\hat{1}_i})$\;
    For the current best alternative $\hat{b}$, set $r_{\hat{b}}^m \gets r_{\hat{b}}^m+1$; then, for each alternative $i \neq \hat{b}$, set $r_i^k \gets r_i^k +1$\;
    \textit{\textbf{$m$-step:}} call $\mathcal{M}$ to allocate $\Delta_m$ among scenarios $\{(\hat{b},1), \dots, (\hat{b}, m)\}$ to get $\left\{\Delta^m_{\hat{b}j}\right\}_{j = 1, \dots, m}$\label{line:$m$-step}\;
    \For{each scenario $(\hat{b}, j)$ with $\Delta^m_{\hat{b}j} > 0$}{
        Take $\Delta^m_{\hat{b}j}$ observations and then update the sample mean $\bar{X}_{\hat{b}j}(n_{\hat{b}j} + \Delta^m_{\hat{b}j})$ and the sample standard deviation $\hat{\sigma}_{\hat{b}j}(n_{\hat{b}j} + \Delta^m_{\hat{b}j})$; set $n_{\hat{b}j} \gets n_{\hat{b}j} + \Delta^m_{\hat{b}j}$ and $n_{\hat{b}j}^m \gets n_{\hat{b}j}^m + \Delta^m_{\hat{b}j}$\;
    }
    \textit{\textbf{$k$-step:}} call $\mathcal{K}$ to allocate $\Delta_k$ among scenarios $\{(i,\hat{1}_i)\}_{i \neq \hat{b}}$ to get $\left\{\Delta^k_{i\hat{1}_i}\right\}_{i \neq \hat{b}}$\label{line:$k$-step}\;
    \For{each scenario $(i,\hat{1}_i)$ with $i \neq \hat{b}$ and $\Delta^k_{i\hat{1}_i} > 0$}{
        Take $\Delta^k_{i\hat{1}_i}$ observations and then update the sample mean $\bar{X}_{i\hat{1}_i}(n_{i\hat{1}_i} + \Delta^k_{i\hat{1}_i})$ and the sample standard deviation $\hat{\sigma}_{i\hat{1}_i}(n_{i\hat{1}_i} + \Delta^k_{i\hat{1}_i})$; set $n_{i\hat{1}_i} \gets n_{i\hat{1}_i} + \Delta^k_{i\hat{1}_i}$ and $n_{i\hat{1}_i}^k \gets n_{i\hat{1}_i}^k + \Delta^k_{i\hat{1}_i}$\;
    }
}
\vspace{6pt}
\KwOut{Alternative $\argmax_{i=1, \dots, k} r_i^m$.}
\end{algorithm} 

Relative to AA, GAA introduces an $m$-step sampling rule $\mathcal M$ with step-wise budget $\Delta_m\geq1$ and a $k$-step sampling rule $\mathcal K$ with step-wise budget $\Delta_k\geq1$. The AA procedure corresponds to equal allocation with $\Delta_m=m$ and $\Delta_k=k-1$. GAA also maintains additional counters: $n_{ij}^m$ and $n_{ij}^k$ record the observations allocated to scenario $(i,j)$ through the two steps, while $r_i^m$ and $r_i^k$ record the number of times alternative $i$ is involved in them. In addition to the sample mean $\bar X_{ij}(n_{ij})$, GAA computes the sample standard deviation $\hat\sigma_{ij}(n_{ij})$ to accommodate variance-aware rules. These statistics require $n_0\geq2$ initial observations per scenario.

In each $m$-step, $\mathcal M$ maps the sample means, sample variances, and sample sizes of the $m$ scenarios of the current best alternative $\hat b$, together with $\Delta_m$, to an allocation $\{\Delta^m_{\hat b j}\}_{j=1,\dots,m}$. Similarly, in each $k$-step, $\mathcal K$ uses the sample information from the $k-1$ current worst-case scenarios $\{(i,\hat 1_i)\}_{i\neq\hat b}$ to produce an allocation $\{\Delta^k_{i\hat 1_i}\}_{i\neq\hat b}$ of $\Delta_k$ observations. For scenarios not involved in a step, we set the corresponding allocation to zero.
When no further complete round fits within the total budget, GAA, like AA, selects the alternative most frequently identified as the current best. Equivalently, it selects the alternative with the largest number of $m$-step observations, $\sum_{j=1}^m n_{ij}^m$. Because $\mathcal M$ and $\mathcal K$ are left unspecified, GAA represents a class of procedures rather than a single procedure.

A readily useful choice for $\mathcal{K}$ and $\mathcal{M}$ is to adopt those embedded in existing fixed-budget procedures developed for traditional R\&S problems. Notable examples include OCBA \citep{chen2000simulation}, KG  \citep{frazier2008knowledge}, TTTS \citep{russo2020simple}, and BOLD \citep{chen2022balancing}, among others. These procedures typically define a sampling rule that, at each round, maps the sample information from all alternatives to an allocation of a batch of $\Delta \geq 1$ observations—precisely the functionality required by $\mathcal{K}$ and $\mathcal{M}$ in the GAA framework. As a result, they can be seamlessly incorporated into GAA.
The rationale extends beyond modular convenience.
A common feature of these R\&S procedures is that they tend to concentrate observations on the best alternative and its close competitors \citep{ryzhov2016convergence, gao2017new, chen2022balancing}. This behavior aligns well with the intended roles of the $k$-step and $m$-step in the AA procedure, as discussed in Section~\ref{subsec: insights}. Intuitively, for the $m$-step—whose purpose is to push up the worst-case sample mean of a non-best alternative that initially appears overly low (i.e., debiasing from below, as illustrated in Figure~\ref{fig:illustration})—it is beneficial to allocate more observations to the scenario with the highest (``best'') true mean among that alternative's $m$ scenarios. 
Conversely,  for the $k$-step—whose purpose is to push down the worst-case sample mean of the true best alternative when it appears overly high (i.e., debiasing from above, as illustrated in Figure~\ref{fig:illustration})—allocating more observations to the true best alternative, which has the lowest worst-case mean across all alternatives, can help hasten this correction. 


\subsection{Implementation}
\label{subsec:implementation}
We would like to highlight that even when the same sampling rule is used for both $\mathcal{M}$ and $\mathcal{K}$, their implementation should differ. As discussed above, $m$-steps are max-seeking while $k$-steps are min-seeking. This interesting directional asymmetry should be carefully respected in implementation. 

Besides, one may use a convenient joint design that unifies the two subproblems by transforming the $k$-step into a max-seeking task. This can be done by taking the negative of the sample means for the $k$-step scenarios. Then, one can concatenate the $k$-step and $m$-step scenarios into a single set. In this set, the current worst-case scenario of the current best alternative can be treated as the best. This transformation allows the use of a single max-seeking sampling rule to allocate the budget across all $k + m - 1$ scenarios. The sequential OCBA procedures in \cite{gao2017robust} and \cite{wan2024new} may intuitively fall into this category.
 Finally, beyond utilizing the sampling rules in existing procedures,  designing new rules tailored specifically for $\mathcal{M}$ and $\mathcal{K}$ in the DRR\&S setting would be interesting, but we leave it for future work.

\subsection{Additive Properties of the General Design}
\label{subsec: general_properties}

The GAA framework allows flexible choices of the \(m\)-step and \(k\)-step sampling rules, $\mathcal{M}$ and $\mathcal{K}$, while preserving the overall structure of the AA procedure. In this subsection, we explore the additivity of the GAA framework. Ideally, we would like to establish analogous results for a broad class of $\mathcal{M}$ and $\mathcal{K}$. To achieve this, we impose the following assumptions on the allocation behavior of $\mathcal{M}$ and $\mathcal{K}$.

\begin{assumption} [Sufficient Exploration in $m$-Steps] 
\label{assu:exploration_m}
Let $\hat{b}(t)$ denote the alternative selected for the $m$-step of the $t$-th round. It holds that almost surely, for every alternative $i=1,\dots, k$, as $t \rightarrow \infty$, if  $r_i^m(t) = \sum_{\tau=1}^t \mathbbm{1}_{ i =  \hat{b}(\tau) } \rightarrow
\infty$, then for each $j=1,\dots,m$, $  n_{ij}^m(t) = \sum_{\tau=1}^t \Delta^m_{ij}(\tau)  \rightarrow \infty$.
\end{assumption}

\begin{assumption}[Sufficient Exploration in $k$-Steps]
\label{assu:exploration_k} 
Let $K(t)=\{1,\dots, k\}\setminus\{ \hat{b}(t)\}$ denote the alternatives selected for the $k$-step of the $t$-th round. It holds that almost surely, $\Delta_{ij}^k(t) \in \{0,1\}$ for all $(i,j)$ and t, and for every alternative $i=1,\dots, k$, as $t \rightarrow \infty$, if $r_i^k(t) =\sum_{\tau=1}^t \mathbbm{1}_{ i \in K(\tau) } \rightarrow
\infty$, then  $  n_i^k(t) = \sum_{\tau=1}^t \sum_{j=1}^m \Delta^k_{ij}(\tau)  \rightarrow \infty$.
\end{assumption}

Assumption~\ref{assu:exploration_m} ensures sufficient exploration in the $m$-steps: it requires that if an alternative enters the $m$-step infinitely often, then each of its scenarios must receive an infinite number of observations through $m$-step allocations. Similarly, Assumption~\ref{assu:exploration_k} imposes sufficient exploration in the $k$-steps: it guarantees that if an alternative appears in the $k$-step infinitely often, its sequence of empirical worst-case scenarios collectively receives infinitely many observations. Intuitively, these exploration requirements correspond to the debiasing roles of the $m$-step and $k$-step, as discussed in Section~\ref{subsec: insights}. The condition $\Delta_{ij}^k(t) \in \{0,1\}$ is introduced purely for technical convenience in establishing the properties of GAA. Importantly, this is not a restrictive assumption: it is naturally satisfied by any fully sequential sampling rule for $\mathcal{K}$, such as KG and TTTS, where only one observation is allocated per $k$-step, i.e., $\Delta_k=1$.

The assumptions are immediate for AA. For adaptive rules, however, they are less direct to verify because the scenarios involved in each step evolve with the coupled sampling process. To rigorously ensure these assumptions, one practical remedy is to incorporate an $\varepsilon$-greedy exploration mechanism (see, e.g., \citealt{li2023convergence}). When invoking $\mathcal{M}$ or $\mathcal{K}$, with a small probability $\varepsilon$ (e.g., 0.1), the step uses a uniform or round-robin allocation across eligible scenarios; with probability $1 - \varepsilon$, it applies the designated rule. 

Under Assumptions~\ref{assu:exploration_m} and \ref{assu:exploration_k}, Theorems~\ref{thm: consistency} and~\ref{thm: additive}, originally established for AA, can be extended to the GAA framework. The result is stated below, with the proof provided in Section~\ref{subsec: proof_GAA_additive}. 

\begin{theorem}[Consistency and Additivity]
    \label{thm: properties_general}
With sampling rules $\mathcal{M}$ and $\mathcal{K}$ satisfying Assumptions~\ref{assu:exploration_m} and \ref{assu:exploration_k}, a GAA procedure is consistent, and it satisfies
\[
\sum_{i=1}^k\sum_{j=1}^m
\mathbbm{1}_{\lim_{N\rightarrow\infty}n_{ij}(N)=\infty}
=k+m-1 \quad \text{almost surely}.
\] 
\end{theorem}

We can also extend Proposition~\ref{prop: possible_limiting} and Theorem~\ref{thm: non_necessity} to GAA. Define $a_{ij}(n_0) := \Phi\left( \sqrt{n_0}(b_\delta - \mu_{ij})/\sigma_{ij} \right)$, and let $b_{ij}$ and $b_{1j}$ be the constants introduced above. Together with Theorem~\ref{thm: properties_general}, this extension provides a characterization of the additive structure for the GAA class. The proof is provided in Section~\ref{subsec: proof_thm_non_necessity_general}.
\begin{proposition}[Possible Limiting Scenarios]
    \label{prop: possible_limiting_general}
For any GAA procedure satisfying Assumptions \ref{assu:exploration_m} and \ref{assu:exploration_k}, fix any non-best alternative $i=2,\dots,k$ and any scenario $j$ such that $\mu_{ij}>\mu_{11}$. For any $b_\delta\in(\mu_{11},\mu_{ij})$,
    \begin{eqnarray*}
         \Pr\left\{\lim_{N\rightarrow\infty}n_{ij}(N)=\infty,\ 
         \lim_{N\rightarrow\infty}n_{i\ell}(N)<\infty,\ \forall\,\ell\neq j\right\} \geq b_{ij}\prod_{\ell\neq j}a_{i\ell}(n_0)\prod_{\ell=1}^{m}b_{1\ell}>0.
    \end{eqnarray*}
\end{proposition}

\begin{theorem}[Non-Necessity of Correct Identification of Worst-Case Scenarios]
\label{thm: non_necessity_general}
For a non-best alternative $i$, if there exists a non-worst-case scenario $j$ such that $\mu_{11}<\mu_{ij}<\mu_{i1}$, then, for any $b_\delta\in(\mu_{11},\mu_{ij})$,
\begin{eqnarray*}
\Pr\left\{\lim_{N\rightarrow\infty}n_{ij}(N)=\infty,\ 
\lim_{N\rightarrow\infty}n_{i\ell}(N)<\infty,\ 
\forall\,\ell\text{ such that }\mu_{i\ell}=\mu_{i1}\right\}
\geq b_{ij}\prod_{\ell\neq j}a_{i\ell}(n_0)\prod_{\ell=1}^{m}b_{1\ell}>0.
\end{eqnarray*}
\end{theorem}

These results show that the additive structure is not exclusive to the simplified AA procedure. Moreover, similar results can be established for the joint design introduced above, which concatenates the $k$-step and $m$-step scenarios, under Assumptions~\ref{assu:exploration_m} and \ref{assu:exploration_k}. In our numerical experiments, we evaluate the performance of several GAA instances, each adopting a different sampling-rule methodology. These instances exhibit budget allocation behaviors similar to those of AA while delivering substantial performance improvements.

\subsection{Toward Asymptotic Budget Allocation Ratios}
\label{subsec: fixed_asymptotic_budget_allocation_ratios}
Theorems~\ref{thm: non_necessity} and \ref{thm: non_necessity_general} show that the GAA procedures need not converge to a predetermined deterministic vector of budget allocation ratios. This contrasts with the prominent large-deviations rate-optimal budget allocation framework. In that framework, the optimal budget allocation ratios are derived by solving an oracle optimization problem that maximizes the PICS convergence rate as $N \to \infty$ \citep{glynn2004large,gao2017new, chen2022balancing}; the resulting solution is deterministic and admits no randomness. For DRR\&S, a natural conjecture is that the optimal ratios would concentrate on the worst-case scenario of each non-best alternative and all scenarios of the best alternative, with all other scenarios receiving a ratio of zero. Our results for sequential designs inspired by common practice follow and go beyond this conjecture: most scenarios receive only a \emph{finite} number of observations, achieving a strong form of additivity. 

Indeed, our results underscore that \emph{fixed-ratio convergence requires additional care in DRR\&S}.
In this paper, we do not enforce convergence to such fixed-ratio structures. Doing so would require correctly identifying the true worst-case scenario of each non-best alternative. Within our framework, one could in principle add a further layer of exploration—for instance, applying a rule such as UCB within the budget that the $k$-step allocates to each non-best alternative to enforce infinite exploration of all its scenarios. However, such enforced exploration would come at the cost of the strongest-sense additivity. We leave characterizing the optimal ratios and designing a procedure that achieves them for future work.
 
A related but distinct open question is how to achieve the optimal PICS convergence rate—noting that tracking the optimal ratios asymptotically does not by itself guarantee it (see, e.g., discussions in \citealt{wu2018analyzing,li2023convergence,chen2022balancing}). We believe this is a genuinely challenging objective, even for traditional R\&S; recently, \cite{goldberger2026fundamental} prove that no algorithm uniformly attains the static-oracle rate across all instances for $k\geq 3$ in one-parameter natural exponential families, a fundamental limit of fixed-budget R\&S. We leave addressing the challenge for DRR\&S for future work. 

\section{Numerical Experiments}
\label{sec: numerical}
In this section, we conduct numerical experiments to verify our theoretical findings and evaluate the proposed procedures. Specifically, Section \ref{subsec:num_properties of AA} illustrates the convergence properties of AA and visualizes its additive sampling behavior. Section \ref{subsec:CRN} investigates the effects of common random numbers (CRNs). Section \ref{sec:properties of GAA} evaluates three GAA procedures, discusses the allocation of $\Delta_m$ and $\Delta_k$, and examines their budget allocation behaviors. Finally, Section \ref{subsec:non-normal} discusses relaxations for the normality assumption and evaluates the GAA procedures in two practically motivated examples with non-normal observations.
 
In the first three subsections, we follow the experimental setups of \cite{fan2020distributionally} and \cite{wan2024new} by considering two illustrative configurations for the true means of the $km$ scenarios:
\begin{itemize}
    \item Slippage configuration (SC): $\mu_{ij} = 0.5 \cdot \mathbbm{1}_{i \neq 1}$, $\forall 1\leq i \leq k, 1\leq j \leq m$.
    \item Monotone means (MM) configuration: 
    $
    \mu_{ij}  = 0.3 (i - 1) - 0.1 (j - 1), \forall 1\leq i \leq k, 1\leq j \leq m.
    $    
\end{itemize}
For both configurations, scenario $(i,1)$ represents the worst-case distribution of alternative $i$, and alternative 1 is the unique best. 
For scenario variances, we adopt a simple constant variance (CV) setting: 
$\sigma_{ij}^2 = 5^2, \forall 1\leq i \leq k, 1\leq j \leq m.$
Together, these define two problem configurations: SC-CV and MM-CV. We consider normally distributed alternatives first. We test problems with non-normal alternatives later in Section \ref{subsec:non-normal}.

In each experiment, the total sampling budget is set as $N = (n_0 + n_1)km$, where $n_0$ denotes the initial sample size per scenario, and $n_1$ controls the number of additional observations allocated in subsequent rounds. Procedure performance is evaluated via the empirical PCS or PICS.
Unless otherwise specified, each PCS or PICS is estimated from $10,000$ independent macro replications.

\subsection{The Simplified AA: Additivity, Convergence of PICS, and the Benefit of Additivity}\label{subsec:num_properties of AA}
\begin{figure}[htp]
    \centering
    \caption{Exponential PICS Decay of the AA Procedure under SC-CV and MM-CV when \(k = 10, m = 5\)}\vspace{5pt}
    \label{fig:decay rate}
    \includegraphics[width=0.7\linewidth]{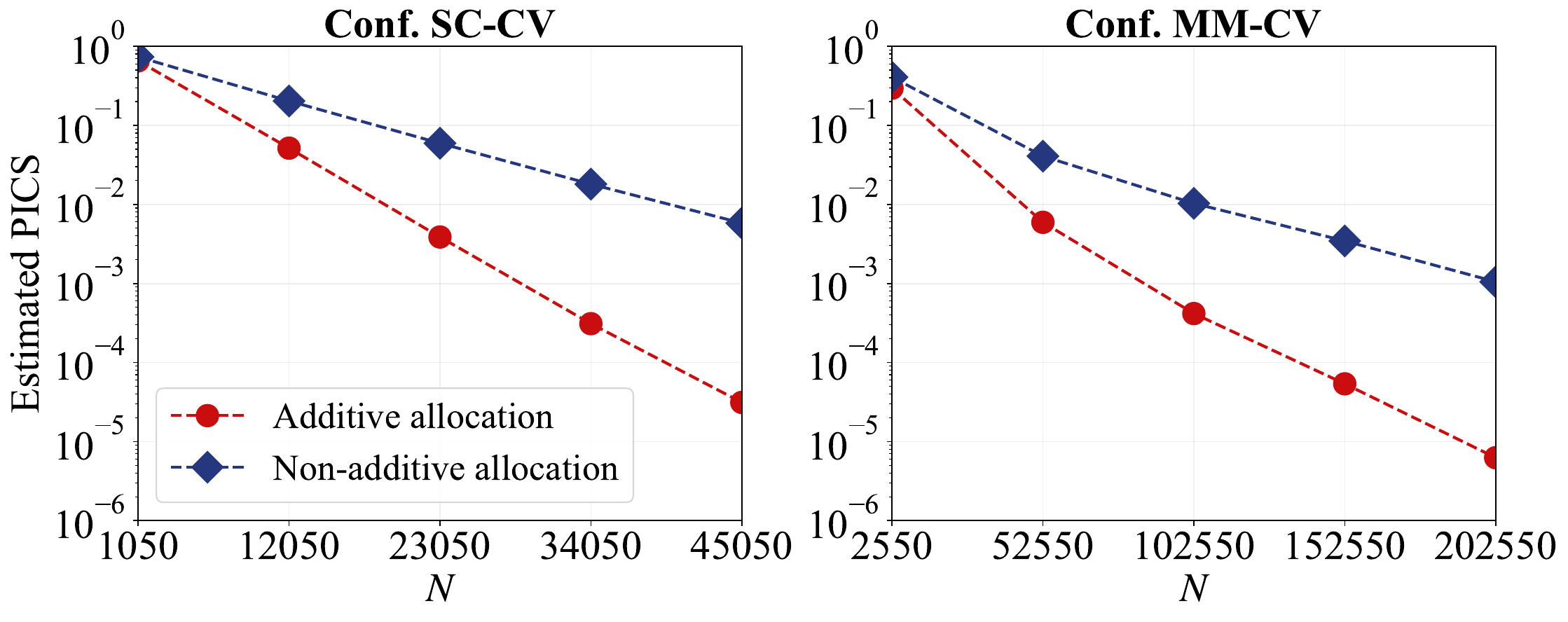}
    \vspace{-15pt}
\end{figure}

We begin by examining the convergence behavior of the PICS for AA as the total sampling budget grows large. We set $k = 10$, $m = 5$, and $n_0 = 1$ in this experiment. Figure~\ref{fig:decay rate} plots the PICS curves of AA under SC-CV and MM-CV, where each data point is based on $800,000$ independent replications. We also include the non-additive equal allocation (EA) procedure as a natural benchmark. 
Figure~\ref{fig:decay rate} reveals two findings. First, the PICS of AA decreases approximately linearly on a logarithmic scale as $N$ increases, consistent with the exponential decay behavior established in Proposition~\ref{prop: exponential_decay}. This trend also provides empirical support for the consistency of AA. Second, although EA also exhibits exponential PICS convergence (due to the normality of alternatives, as in traditional R\&S problems), AA achieves a substantially faster decay under both configurations. Note that the main difference between AA and EA is that AA tries to be additive in each round. The comparison therefore provides empirical evidence for the benefit of the additive design.

\begin{figure}[htp]
\vspace{-5pt} 
     \FIGURE
    {\includegraphics[width=0.95\linewidth]{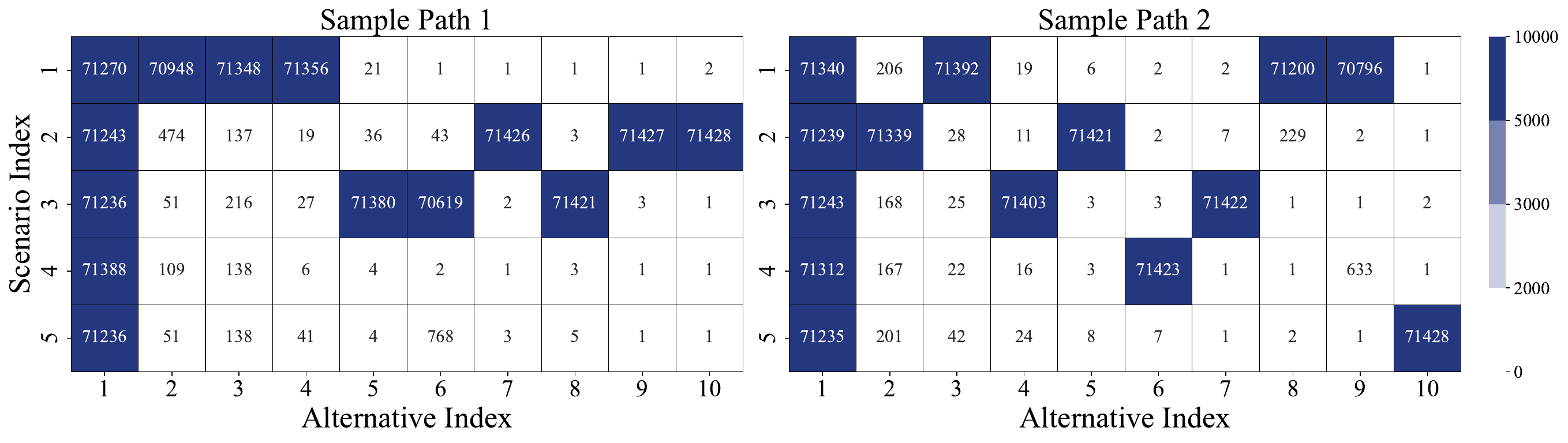}}
     {Sample Allocation Pattern of the AA Procedure under MM-CV when \(k = 10, m = 5\)    \vspace{5pt}\label{fig: sample allocation of AA}}
    {Each panel shows the final sample sizes across all $k \times m = 50$ scenarios for one sample path.}
    \vspace{-10pt}
\end{figure}

To further examine AA's budget allocation behavior, 
Figure~\ref{fig: sample allocation of AA} displays the final sample sizes across all scenarios under the MM-CV configuration, using a total budget of $N = 20,000 km$ ($n_0=1$ and $n_1 = 19,999$). We present two sample paths in which a correct selection is achieved. The results under SC-CV are similar and are omitted. For the same reason, the experiments that follow also report only the MM-CV results. Across both paths, the results exhibit the additive structure predicted by Theorem~\ref{thm: additive}. Among the $k \times m = 50$ scenarios, only $k + m - 1 = 14$ scenarios receive a substantial number of observations: these include all scenarios of the best alternative, as well as exactly one scenario from each non-best alternative. The remaining scenarios receive only a negligible sampling effort.

Furthermore, as predicted by Theorem \ref{thm: non_necessity}, the most heavily sampled scenario within each non-best alternative --- interpreted as the estimated critical scenario --- does not always coincide with the true worst-case scenario and varies across runs. For example, the most sampled scenario for alternative 5 is $(5,3)$ in sample path 1 and $(5,2)$ in sample path 2 --- neither of which corresponds to the true worst-case scenario (5,1). Despite this mismatch, AA still achieves the convergence properties of the PICS demonstrated in Figure \ref{fig:decay rate}.

\subsection{Effects of CRNs}\label{subsec:CRN}
Next, we relax the independence assumption to investigate the effect of CRNs.  
Here, the two-layer structure of DRR\&S makes the use of CRNs less straightforward than in traditional R\&S: the comparisons arise in two different dimensions, and CRNs may be introduced either \emph{among scenarios within the same alternative} or \emph{among alternatives under the same input distribution}. 
To investigate the effect of each dimension, we control the dependence among the underlying standard normal inputs under MM-CV using the parameter $\rho \in [0,1]$. These inputs are then scaled and shifted to generate normally distributed observations. When $\rho = 0$, the relevant scenarios are mutually independent. When $\rho = 1$, the relevant scenarios use identical standard normal inputs.
For any two scenarios \((i,j)\) and \((i',j')\), with indexed outputs \(X_{ijp}\) and \(X_{i'j'q}\), dependence is introduced only when \(p=q\), while outputs from different replications (\(p\neq q\)) remain independent.

\begin{figure}[tp]
    \FIGURE
   {\includegraphics[width=0.7\linewidth]{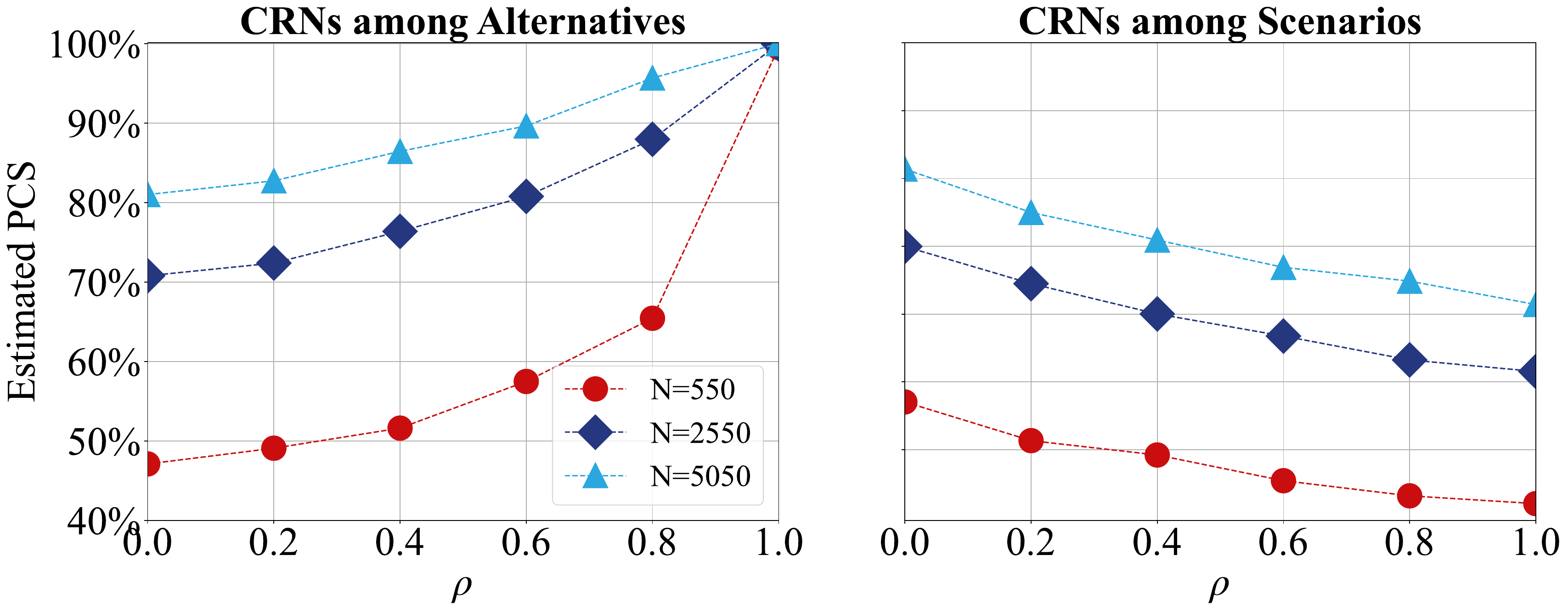}}
    {Effects of CRNs on the PCS of the AA Procedure under MM-CV when \(k = 10, m = 5\) \vspace{5pt}    \label{fig:CRN}}
   {The left panel applies CRNs across alternatives evaluated under the same input distribution, whereas the right panel applies CRNs across scenarios within the same alternative. In each case, random inputs remain independent along the other dimension.}
\vspace{-10pt}
\end{figure}

Figure \ref{fig:CRN} reveals a sharp contrast between two implementations of CRNs. 
The left panel considers correlation among alternatives under the same input distribution. The PCS increases steadily with $\rho$ under all three budget levels. Under this implementation, alternatives evaluated under the same input distribution are exposed to similar random variation. Intuitively, as in traditional R\&S, this makes comparisons across alternatives more reliable.
The right panel considers correlation among scenarios within the same alternative and, interestingly, shows the opposite pattern: the PCS decreases as $\rho$ increases. To understand this result, suppose that a non-best alternative receives an unusually favorable (i.e., low) random input. When its scenarios are strongly positively correlated, their observations tend to move downward together. Consequently, all its scenario sample means, including its empirical worst-case mean, may appear better than they should, increasing the chance of being selected as the current best. Under weaker correlation, the scenarios of a non-best alternative exhibit more independent variation. Even if some scenarios appear favorable, another scenario is more likely to reveal that the alternative is inferior. 
These findings suggest that the use of CRNs should account for the two-layer structure rather than treating CRNs as a universally effective device.
 
\subsection{Evaluation of GAA Procedures, New and Old} 
\label{sec:properties of GAA}

We now examine the performance and budget allocation behaviors of GAA procedures. We consider three sampling rules: TTTS, KG, and BOLD. GAA-KG and GAA-BOLD apply KG and BOLD, respectively, to both the $m$-step and $k$-step. We also consider a joint design: GAA-TTTS applies TTTS to the combined set of $m$-step and $k$-step scenarios, as described in Section \ref{subsec:implementation}. By default, $\Delta_m=1$ and $\Delta_k=1$.  More implementation details for the GAA procedures are provided at \href{https://github.com/Wanone10001/DRRnS}{https://github.com/Wanone10001/DRRnS} due to space limit. 
We do not consider new OCBA-based variants, as the OCBA approach has already been adopted in the design of DRR\&S procedures \citep{gao2017robust,wan2024new}. They intuitively can be treated as instances of the GAA framework applying OCBA rules jointly to the combined set of $m$-step and $k$-step. Thus, their sampling behaviors may be understood under the same framework; see Section \ref{sec:existingsamplepath} for more illustration. We include R-OCBA \citep{gao2017robust} as an OCBA instance. We also include the non-additive EA procedure as a benchmark and R-UCB \citep{wan2023upper} for comparison.
Here we set the initial sample size to $n_0 = 20$ to facilitate initial variance estimation.

\subsubsection{Performance Comparison.}

\begin{figure}[pt]
    \vspace{-5pt}
    \FIGURE
   {\includegraphics[width=0.95\linewidth]{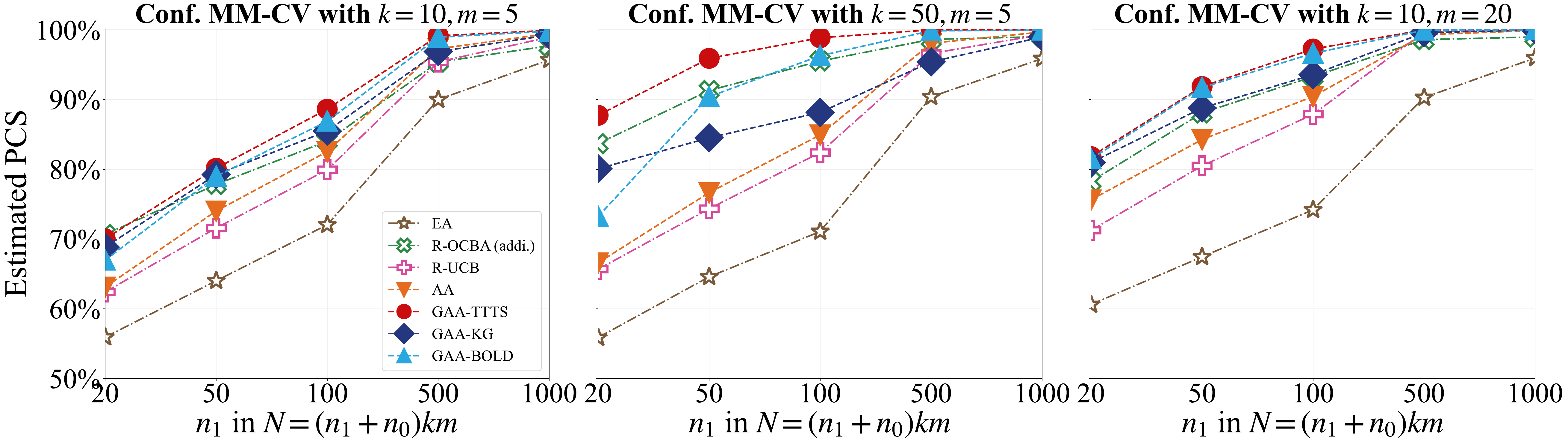}}
    {Comparative Performance of GAA Procedures with Different Sampling Rules under MM-CV \vspace{5pt}    \label{fig:consistency of GAA}}
{}
\vspace{-10pt}
\end{figure}

Figure \ref{fig:consistency of GAA} shows the PCS curves of each procedure as the sampling budget grows for each problem scale. We can observe that GAA procedures (including R-OCBA) achieve consistently better performance than the others. The gap between AA and GAA procedures showcases the value of using adaptive sampling rules.
Also, consistent with Theorem \ref{thm: properties_general}, their PCS approaches near 100\% as the budget increases, indicating consistency. Among them, GAA-TTTS achieves the highest PCS in nearly all settings; this is not surprising, as TTTS has emerged as a popular rule in various learning problems due to its strong performance. GAA-BOLD also performs strongly and is generally close to GAA-TTTS, while GAA-KG remains competitive with the benchmarks. These results may show the practical value of our analysis, which not only provides structural understanding but also yields new effective procedures.

\subsubsection{Additivity.} 
\label{subsubsec: detail_GAA_exp}
\begin{figure}[htbp]
    \vspace{-5pt}
     \FIGURE
    {\includegraphics[width=0.95\linewidth]{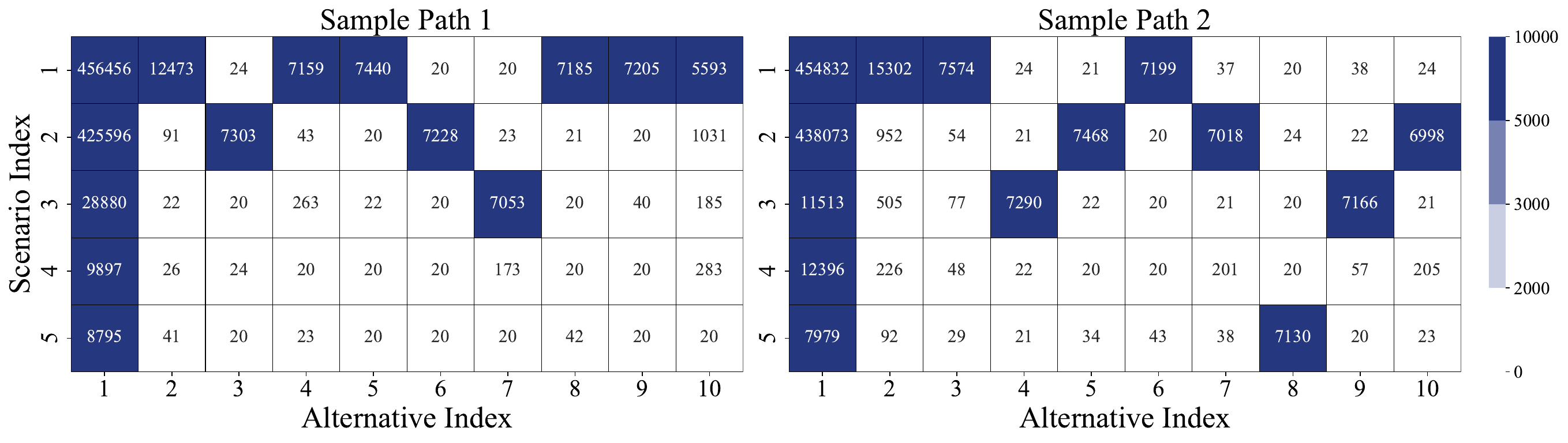}}
     {Sample Allocation Pattern of GAA-TTTS under MM-CV when \(k = 10, m = 5\) \vspace{5pt}    \label{fig: allocation of GAA}}
    {}
    \vspace{-15pt}
\end{figure}

Using the same visualization setup as in Section~\ref{subsec:num_properties of AA} (except setting $n_0 = 20$), we display the final sample allocations in two representative sample paths (where a correct selection is achieved) of GAA-TTTS under the MM-CV configuration in Figure~\ref{fig: allocation of GAA}. The other GAA procedures exhibit similar allocation patterns and are thus omitted. Similar to the behavior observed for AA, GAA-TTTS concentrates most of the budget on $k + m - 1$ scenarios, indicating additivity. Also, the most heavily sampled scenarios in non-best alternatives are not always the true worst-case scenarios, reinforcing the observation that exact worst-case identification may be unnecessary, as predicted by Theorem~\ref{thm: non_necessity_general}.

\subsubsection{Determining $\Delta_m$ and $\Delta_k$.}
\label{subsubsec:delta_allocation} 
\begin{figure}[htp]
    \vspace{-5pt}
    \FIGURE
   {\includegraphics[width=0.7\linewidth]{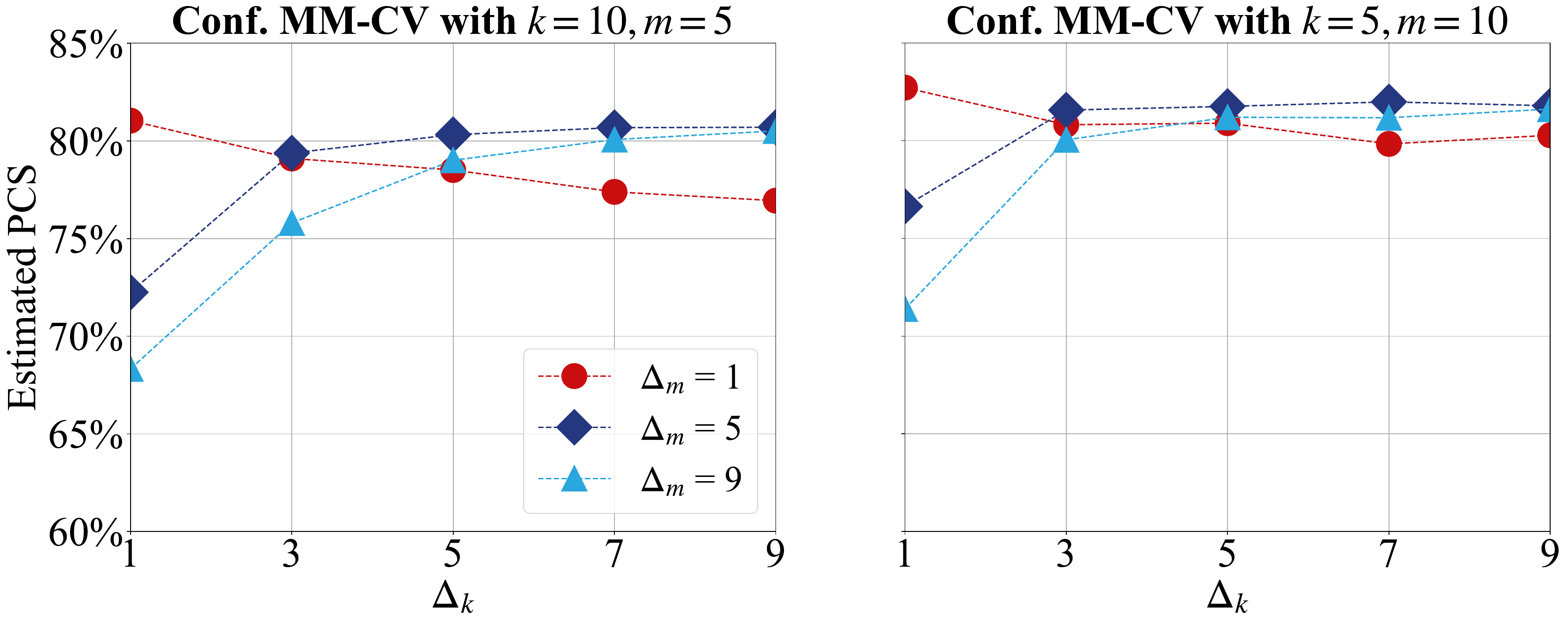}}
    {PCS of GAA-KG under Different $\Delta_m$ and $\Delta_k$\vspace{5pt}    \label{fig:delta_allocation}}
   {}
    \vspace{-15pt}
\end{figure}

$\Delta_m$ and $\Delta_k$ are two important parameters in the GAA framework. We now examine how their choices affect the performance of GAA. Here we use GAA-KG as a representative example.
We vary $\Delta_m \in \{1,5,9\}$ and $\Delta_k \in \{1,3,5,7,9\}$ independently, while the overall sampling budget remains fixed ($n_0 = 20$ and $n_1 = 60$). Within each step, the assigned observations are allocated sequentially, and the KG decision is updated after each observation. We consider two problem scales representing settings with more alternatives than scenarios and more scenarios than alternatives, respectively. 
Figure \ref{fig:delta_allocation} shows that $(\Delta_m, \Delta_k) = (1,1)$ achieves the highest PCS under both problem scales, which is not surprising, as smaller $\Delta_m, \Delta_k$ means more additive rounds and more opportunities to revise the current worst-case scenarios under a fixed total budget. We also observe that $(\Delta_m, \Delta_k)$ should be kept at a comparable scale.
When $\Delta_m=1$ or $ \Delta_k=1$, increasing the other one is not beneficial. When $\Delta_m$ is moderate (5 and 9), it is helpful to increase $\Delta_k$. This phenomenon can be understood from the complementary roles of the $m$-step and $k$-step, as illustrated in Section \ref{subsec: insights}. Based on our experimental findings, we recommend $(\Delta_m, \Delta_k) = (1,1)$; otherwise, if fewer rounds are preferred, one should keep $(\Delta_m, \Delta_k)$ balanced.

\subsection{Problems with Non-Normal Observations}
\label{subsec:non-normal}
We adopt the  normality assumption throughout the paper for ease of presentation and analysis. However, the proposed framework is not limited to normal distributions. Algorithmically, GAA procedures can adapt to specific distributional settings, as the sampling rules $\mathcal{M}$ and $\mathcal{K}$ are customizable. The additive structure holds quite generally. Our boundary-crossing analysis in Section \ref{subsec:boundary}, which is key to understanding additivity, would require only continuity of the cdf to avoid the event of the sample mean hitting the boundary, as mentioned below Observation \ref{obs: non-best}. One could also relax this assumption with additional care.  The exponential decay of PICS in Proposition \ref{prop: exponential_decay}, however, requires a light-tailed assumption on the distribution. We numerically verify this for the exponential distribution (result not shown).
In the following, we demonstrate the applicability of GAA to problems with non-normal observations through two practically motivated examples. We briefly describe each problem and summarize the results below. Full implementation details and parameter settings are provided at \href{https://github.com/Wanone10001/DRRnS}{https://github.com/Wanone10001/DRRnS} due to space limit. We compare the additive procedures GAA-TTTS, GAA-KG, GAA-BOLD, and R-OCBA with EA and R-UCB.

\textbf{Inventory Management:} In this example, each alternative corresponds to an $(s, S)$ inventory policy, where $s$ denotes the reorder point and $S$ the order-up-to level. The set of alternatives is constructed by enumerating combinations of $s \in \{240, 260, 280, 300, 320, 340\}$ and $S \in \{350, 370, 390, 410, 430, 450\}$. Customer demand uncertainty is modeled through an ambiguity set of exponential distributions with mean $\mu \in \{310, 320, 330, 340\}$, while customer arrivals follow a known Poisson process. The objective is to select the $(s, S)$ policy that minimizes long-run average cost, under demand uncertainty. 

\textbf{Multiserver Queuing:}
This example, adapted from \cite{fan2020distributionally}, considers the problem of determining the optimal number of servers in a multiserver queue with customer abandonment, aiming to minimize the total cost of staffing, customer waiting, and abandonment. Each alternative corresponds to a staffing level chosen from $\{9, 10, 11, 12\}$. Service time distribution uncertainty is captured via an ambiguity set constructed by fitting candidate distributions (log-normal, gamma, Weibull, and exponential) to a set of 20 empirical observations generated from the true distribution, based on the Kolmogorov–Smirnov goodness-of-fit test.

\begin{figure}[htp]\vspace{-5pt}
    \centering
    \caption{PCS Comparison under Inventory Management and Multiserver Queuing Problems} \vspace{5pt}
    \includegraphics[width=0.8\linewidth]{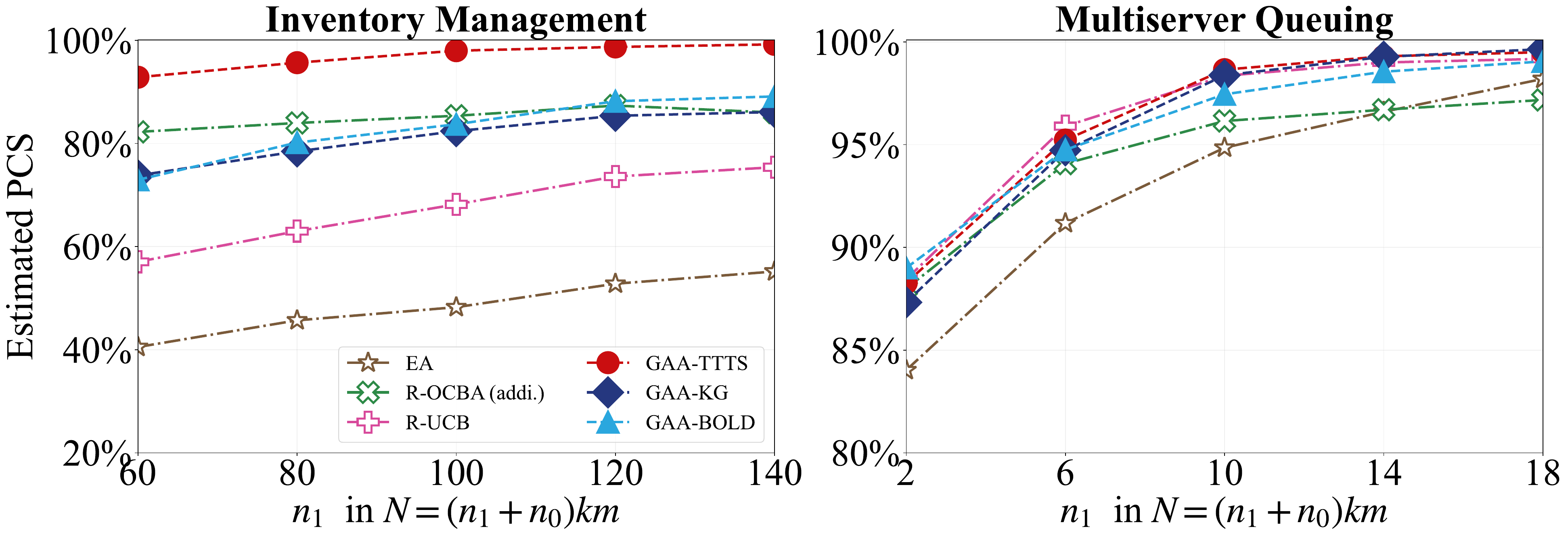}
    \label{fig:examples}
    \vspace{-10pt}
\end{figure}

Across both examples, the GAA procedures perform strongly, with GAA-TTTS consistently achieving the highest PCS, demonstrating a similar pattern under normal observations (Section~\ref{sec:properties of GAA}). In the second example, R-OCBA may underperform R-UCB; however, new GAA procedures may still achieve comparable performance. These results again showcase the potential benefits of the GAA framework. 

\section{Conclusion}
\label{sec: conclusion}
This paper studies DRR\&S through the lens of sequential additivity, advancing the understanding of additivity behind robust R\&S. We first establish an algorithm-independent necessary sampling condition, reflecting a fundamental asymmetry in the problem. We then abstract a common sequential-heuristic design into a simple AA procedure, which helps explicate the complementary rules of the $m$-steps and $k$-steps. Using boundary-crossing arguments, we establish its consistency with an exponential PICS decay rate. We then prove that AA attains the sampling lower bound exactly, showing that additivity can be achieved in the strongest possible sense. Interestingly, the scenario sampled infinitely often for a non-best alternative need not be its true worst-case scenario, indicating that correct worst-case identification may not be necessary in DRR\&S. Building on these structural insights, we develop the GAA framework, which may retain the key properties of AA. This generalizes the additivity structure while offering improved practical performance.

Several directions remain for future research, in addition to those discussed in Section~\ref{subsec: fixed_asymptotic_budget_allocation_ratios}. First, extending the framework beyond finite categorical ambiguity sets to structured or continuous ambiguity sets would introduce important analytical and computational challenges. Second, AA and GAA are currently developed under a sequential computing paradigm; adapting their additive structure to parallel simulation is practically important. 
Finally, the additive structure studied here may extend beyond DRR\&S. Selection problems based on pairwise comparisons and dueling-bandit problems, for example, also possess two-layer minimax structures. Similar forms of sequential additivity may therefore arise in these settings.

\bibliographystyle{informs2014}
\bibliography{ref.bib}
\OneAndAHalfSpacedXI

\ECSwitch
\EquationsNumberedBySection
\vspace{-26pt}
\SingleSpacedXI
\begin{APPENDICES}

\section{Technical Supplement to Section \ref{sec: problem}}

\subsection{Proof of Theorem~\ref{thm:necessary-sampling}}
\label{subsec:sampling_lower_bound}
\subsubsection{Preliminaries.} 
We use \(t\) to index individual sampling times here. Let \(A_t=(i,j)\) denote the scenario sampled at time \(t\), and let \(X_t\) denote the corresponding simulation output. Let \(\mathcal F_t\) contain the sampled scenarios, observed outputs, and possible internal randomization revealed through time \(t\). Any randomization used to form the final selection \(\hat b_N\) is also included in \(\mathcal F_N\), so that \(\{\hat b_N=i\}\in\mathcal F_N\). Conditionally on \(\mathcal F_{t-1}\), let \(Q_t((i,j)\mid\mathcal F_{t-1})\) denote the probability that scenario \((i,j)\) is sampled at time \(t\). Because the procedure is nonanticipating and does not depend on unknown problem parameters, the same sampling kernels are used under every problem instance.

We first establish a finite-sampling transfer result. The idea behind the lemma is that if the scenarios distinguishing two instances are sampled only finitely often, the procedure cannot accumulate unlimited statistical evidence separating those instances. On sample paths where the relevant observations are finite in number and moderate in value, the likelihood ratio remains bounded away from zero. Consequently, an event that becomes certain under one instance must retain positive probability under the other.

\begin{lemma}[Finite-Sampling Transfer]
    \label{lem:finite-sampling-transfer}
    Let \(\mathcal S\) be a set of scenarios, and consider two mean matrices \(\mu\) and \(\mu'\) that differ only on \(\mathcal S\), with all variances unchanged. Define \(n_{\mathcal S}(N)=\sum_{(i,j)\in\mathcal S}n_{ij}(N)\) and \(n_{\mathcal S}(\infty)=\lim_{N\to\infty}n_{\mathcal S}(N)\). Suppose that, for some finite \(M\), \(p:=\Pr_{\mu}\{n_{\mathcal S}(\infty)\leq M\}>0\). If \(B_N\in\mathcal F_N\) and \(\Pr_{\mu}\{B_N\}\to1\), then \(\liminf_{N\to\infty}\Pr_{\mu'}\{B_N\}>0\).
    \end{lemma}
    
    \begin{proof}{Proof.}
    Let \(f_{ij}^{\mu}\) and \(f_{ij}^{\mu'}\) denote the densities of \(\mathcal N(\mu_{ij},\sigma_{ij}^2)\) and \(\mathcal N(\mu'_{ij},\sigma_{ij}^2)\), respectively. Define \(g_{ij}(x)=f_{ij}^{\mu'}(x)/f_{ij}^{\mu}(x)\) and \(L_N=\prod_{t=1}^N g_{A_t}(X_t)\). Because \(\mu\) and \(\mu'\) differ only on \(\mathcal S\), \(g_{ij}(x)=1\) for \((i,j)\notin\mathcal S\). Moreover, because Gaussian densities are everywhere positive, the finite-horizon distributions under \(\mu\) and \(\mu'\) are mutually absolutely continuous.
    
    Under either instance, the density of the observed history factors into the sampling kernels and the corresponding observation densities. Because the same sampling kernels are used under \(\mu\) and \(\mu'\), they cancel from the likelihood ratio. Consequently,
    \[
    L_N
    =
    \frac{d(\Pr_{\mu'}|_{\mathcal F_N})}
    {d(\Pr_{\mu}|_{\mathcal F_N})},
    \qquad
    \Pr_{\mu'}\{D\}
    =
    \E_{\mu}[L_N\mathbbm 1_D],
    \quad \forall D\in\mathcal F_N.
    \]
    
    Let \(E_M=\{n_{\mathcal S}(\infty)\leq M\}\), so that \(\Pr_{\mu}\{E_M\}=p\). Choose \(C<\infty\) sufficiently large that \(M\max_{(i,j)\in\mathcal S}\Pr_{\mu}\{|X_{ij}|>C\}<p/4\). Define \(G_N\) as the event that every observation obtained from \(\mathcal S\) by time \(N\) lies in \([-C,C]\).
    To account for adaptive sampling, let \(\tau_r\) denote the sampling time at which the \(r\)th observation from \(\mathcal S\) is obtained. On \(E_M\cap G_N^c\), at least one of the first \(M\) observations from \(\mathcal S\) lies outside \([-C,C]\). Conditional on \(\mathcal F_{\tau_r-1}\) and the selected scenario \(A_{\tau_r}\), the observation \(X_{\tau_r}\) has the corresponding Gaussian distribution under \(\mu\). Therefore, each term in the union bound is bounded by the largest scenario-wise tail probability over \(\mathcal S\), giving
    \[
    \Pr_{\mu}\{E_M\cap G_N^c\}
    \leq
    \sum_{r=1}^M
    \Pr_{\mu}\{\tau_r\leq N,\ |X_{\tau_r}|>C\}
    \leq
    M\max_{(i,j)\in\mathcal S}
    \Pr_{\mu}\{|X_{ij}|>C\}
    <
    \frac{p}{4}.
    \]
    
    Since \(\Pr_{\mu}\{B_N^c\}\to0\), for all sufficiently large \(N\), \(\Pr_{\mu}\{B_N^c\}<p/4\). A technical complication is that \(E_M\) depends on the entire infinite sampling path and is therefore not necessarily \(\mathcal F_N\)-measurable. We address this by replacing it with the finite-time event \(D_N\), while using \(E_M\) only to lower-bound the probability of \(D_N\). Specifically, define \(D_N=B_N\cap\{n_{\mathcal S}(N)\leq M\}\cap G_N\). Because \(E_M\subseteq\{n_{\mathcal S}(N)\leq M\}\), for all sufficiently large \(N\),
    \[
    \Pr_{\mu}\{D_N\}
    \geq
    \Pr_{\mu}\{E_M\cap B_N\cap G_N\}
    \geq
    p-\Pr_{\mu}\{B_N^c\}-\Pr_{\mu}\{E_M\cap G_N^c\}
    \geq
    \frac{p}{2}.
    \]
    
    The Gaussian likelihood ratios are continuous and strictly positive. Therefore, \(K_C:=\max_{(i,j)\in\mathcal S,\,|x|\leq C}|\log g_{ij}(x)|<\infty\). On \(D_N\), at most \(M\) observations are obtained from \(\mathcal S\), and all of them lie in \([-C,C]\). Hence, \(\log L_N\geq-MK_C\), so \(L_N\geq\exp(-MK_C)\) on \(D_N\). It follows that
    \[
    \Pr_{\mu'}\{B_N\}
    \geq
    \Pr_{\mu'}\{D_N\}
    =
    \E_{\mu}[L_N\mathbbm 1_{D_N}]
    \geq
    \exp(-MK_C)\Pr_{\mu}\{D_N\}
    \geq
    \frac{p}{2}\exp(-MK_C)>0.
    \]
    Therefore, \(\liminf_{N\to\infty}\Pr_{\mu'}\{B_N\}\geq(p/2)\exp(-MK_C)>0\), which proves the lemma. \hfill\Halmos
    \end{proof}

\subsubsection{Proof of Theorem~\ref{thm:necessary-sampling}.} We now use Lemma~\ref{lem:finite-sampling-transfer} to prove the necessary sampling condition.

\begin{proof}{Proof of Theorem~\ref{thm:necessary-sampling}.}
    Fix a problem instance satisfying the conditions of the theorem. Thus, alternative \(1\) is uniquely best and alternative \(2\) is uniquely second-best, with \(\mu_{11}<\mu_{21}<\mu_{i1}\) for \(i=3,\ldots,k\).
    We apply the transfer lemma through two complementary perturbations. 
    For a scenario of the best alternative, we increase its mean until the best alternative becomes nonoptimal. For a non-best alternative, we decrease all its scenario means until it becomes uniquely optimal. 
    Consistency must distinguish the original instance from both types of perturbations.
    
    We first consider the scenarios of alternative \(1\). Fix \(j_0\in\{1,\ldots,m\}\) and suppose, toward a contradiction, that \(\Pr_{\mu}\{n_{1j_0}(\infty)<\infty\}>0\). Because \(n_{1j_0}(\infty)\) is integer-valued whenever it is finite, there exists \(M<\infty\) such that \(\Pr_{\mu}\{n_{1j_0}(\infty)\leq M\}>0\).
    Construct another instance \(\mu'\) by increasing only \(\mu_{1j_0}\) so that \(\mu'_{1j_0}>\mu_{21}\), while leaving all other means and variances unchanged. Under \(\mu'\), the robust value of alternative \(1\) exceeds \(\mu_{21}\). Because alternative \(2\) is uniquely second-best under \(\mu\), it becomes uniquely best under \(\mu'\).
    
    Let \(B_N=\{\hat b_N=1\}\). Consistency under \(\mu\) gives \(\Pr_{\mu}\{B_N\}\to1\). Applying Lemma~\ref{lem:finite-sampling-transfer} with \(\mathcal S=\{(1,j_0)\}\) gives \(\liminf_{N\to\infty}\Pr_{\mu'}\{\hat b_N=1\}>0\). However, because alternative \(2\) is uniquely best under \(\mu'\), consistency requires \(\Pr_{\mu'}\{\hat b_N=1\}\to0\), which is a contradiction. Thus, \(\Pr_{\mu}\{n_{1j}(\infty)=\infty\}=1\) for every \(j=1,\ldots,m\). Because \(m\) is finite, these statements hold simultaneously almost surely.
    
    We next consider a non-best alternative \(i\geq2\). Suppose, toward a contradiction, that \(\Pr_{\mu}\{\sum_{j=1}^m n_{ij}(\infty)<\infty\}>0\). Then, there exists \(M<\infty\) such that \(\Pr_{\mu}\{\sum_{j=1}^m n_{ij}(\infty)\leq M\}>0\).
    For any \(\epsilon>0\), construct \(\mu'\) by setting \(\mu'_{ij}=\mu_{11}-\epsilon\) for \(j=1,\ldots,m\), while leaving all other means and variances unchanged. Alternative \(i\) is then uniquely best under \(\mu'\), because its robust value is \(\mu_{11}-\epsilon\), whereas every other alternative has robust value at least \(\mu_{11}\).
    
    Apply Lemma~\ref{lem:finite-sampling-transfer} with \(\mathcal S=\{(i,1),\ldots,(i,m)\}\) and \(B_N=\{\hat b_N=1\}\). Consistency under \(\mu\) gives \(\Pr_{\mu}\{B_N\}\to1\), whereas the lemma gives \(\liminf_{N\to\infty}\Pr_{\mu'}\{\hat b_N=1\}>0\). This contradicts consistency under \(\mu'\), which requires \(\Pr_{\mu'}\{\hat b_N=1\}\to0\) because alternative \(i\) is uniquely best. Therefore, \(\Pr_{\mu}\{\sum_{j=1}^m n_{ij}(\infty)=\infty\}=1\) for every \(i=2,\ldots,k\).
    

    The finiteness of \(k\) and \(m\) implies that all these conditions hold simultaneously almost surely. Hence, all \(m\) scenarios of the best alternative and at least one scenario of every non-best alternative are sampled infinitely often. 
    \hfill\Halmos
    \end{proof}

\section{Technical Supplement to Section \ref{sec:AA property}}

\subsection{Proof of Lemma \ref{lem: lastexit_tail}}
\label{subsec: proof_lemma_lastexit_tail}
\begin{proof}{Proof.}
Notice that by the symmetry of the standard normal distribution around 0, $U(b)$ and $L(-b)$ are identically distributed. So it suffices to prove the result for $U(b)$. By the definition of $U(b)$, we have that for any $b>0$ and $n \in \mathbbm{N}^{+}$,
\begin{eqnarray}
\label{eq: bound1}
    \notag \Pr \left\{U(b) > n\right\} =
   \Pr \left\{\sup\{n' \geq 1: \bar Z(n') \geq b\} > n\right\}
    &=& 1- \Pr \left\{\sup\{n' \geq 1: \bar Z(n') \geq b\} \leq n\right\} \\
   &=& 1- \Pr \left\{ \max_{n' \geq n+1} \bar Z(n') < b\right\} \leq 1- \Pr \left\{ \max_{n' \geq n} \bar Z(n') < b\right\}.\quad \quad \quad 
\end{eqnarray}
Then, again, by the symmetry of the standard normal distribution around 0, we have 
\begin{eqnarray}
    \label{eq: bound2}
    \Pr \left\{ \max_{n' \geq n} \bar Z(n') < b\right\} = \Pr \left\{ \min_{n' \geq n} \bar Z(n') > -b\right\},
\end{eqnarray}
which, by Lemma 3 of \cite{itemLi2024arxiv}, satisfies 
\begin{eqnarray}
    \label{eq: bound3}
    \Pr \left\{ \min_{n' \geq n} \bar Z(n') > -b\right\} \geq 1-2 \exp\left( - \frac{n b^2}{2} \right).
\end{eqnarray}
Combining Equations \eqref{eq: bound3}, \eqref{eq: bound2}, and \eqref{eq: bound1} leads to result of interest.
\hfill \Halmos
\end{proof}

\subsection{Proof of Lemma \ref{lem: finite}}
\label{subsec: proof_lemma_finite}
\begin{proof}{Proof.}
As in the proof of Lemma~\ref{lem: lastexit_tail}, we prove the result for $U(b)$. For any $b > 0$, let $A_n := \left\{ U(b) > n \right\}$. Then $\left\{ U(b) = \infty \right\} = \bigcap_{n \ge 1} A_n$, and the sequence $\left\{ A_n \right\}_{n \ge 1}$ is decreasing. Furthermore, by Lemma~\ref{lem: lastexit_tail}, we have
$$
\sum_{n=1}^\infty \Pr\left\{ A_n \right \}  \le  \sum_{n=1}^\infty 2\exp\left( - \frac{n b^2}{2} \right) <  \infty.
$$
Then, by the (first) Borel–Cantelli lemma (which does not require independence), we have $\Pr\left\{ A_n\ \text{i.o.} \right\} = 0$. Since $\left\{ A_n \right\}_{n \ge 1}$ is decreasing, we also have $\left\{ A_n\ \text{i.o.} \right\} = \bigcap_{n \geq 1} A_n = \left\{ U(b) = \infty \right\}$. Therefore, $\Pr  \left\{ U(b) = \infty \right\} = 0$. The same argument applies to $L(-b)$, which concludes the proof of Lemma~\ref{lem: finite}. \hfill \Halmos
\end{proof}

\subsection{Proof of Proposition \ref{prop: exponential_decay}}
\label{subsec: proof_exponential_decay}
\begin{proof}{Proof.} Let $q=m+k-1$, $T=\lfloor(N-mk)/q\rfloor$, and $r=\max\{0,\lceil T/(2q)\rceil-1\}$. When $T=0$, the stated bound is trivial because $r=0$ and its right-hand side exceeds one. Suppose that $T\geq1$. From the PCS lower bound in Lemma~\ref{lem: PCS-bound}, the PICS satisfies
\begin{eqnarray*}
\mathrm{PICS}
\leq
\Pr\left\{T\leq2S(b_\delta)\right\}
=
\Pr\left\{S(b_\delta)\geq T/2\right\}.
\end{eqnarray*}
Because $r=\lceil T/(2q)\rceil-1<T/(2q)$, we have $qr<T/2$. Thus, if $S(b_\delta)\geq T/2$, at least one of its $q$ nonnegative terms must exceed $r$. The union bound therefore gives
\begin{eqnarray}
\label{eq: twoparts}
\mathrm{PICS}
\leq\sum_{i=2}^k \Pr\left\{ \min_{j=1,\dots,m} L_{ij}(b_\delta) > r\right\}
+ \sum_{j=1}^m \Pr\left\{ U_{1j}(b_\delta) > r\right\}.
\end{eqnarray}

We now analyze the first term. As highlighted in Section~\ref{subsubsec:last_exit_times}, the last exit time $L_{ij}(b_\delta)$ is defined with respect to the entire sample mean process $\{\bar{X}_{ij}(n)\}_{n=1}^\infty$, from $n = 1$ to $\infty$, and not based on the sampling process of the procedure. Since observations are independent across scenarios, $L_{ij}(b_\delta)$ are mutually independent across scenarios $j$. Therefore,
\begin{eqnarray}
\label{eq:productbound}
    \Pr\left\{ \min_{j=1,\dots,m} L_{ij}(b_\delta) > r \right\} = \Pr\left\{\forall j=1,\dots,m: L_{ij}(b_\delta) > r \right\}= \prod_{j=1}^m \Pr\left\{ L_{ij}(b_\delta) > r \right\}. 
\end{eqnarray}
When $\mu_{ij} > b_\delta$, Lemma~\ref{lem: lastexit_tail} gives, for $r\geq1$,
\begin{eqnarray}
\label{eq:meanhigher}
    \notag \Pr\left\{ L_{ij}(b_\delta) > r \right\} & = &  \Pr\left\{ \sup \{ n \geq 1 : \bar{X}_{ij}(n) \leq b_\delta \} > r \right\}  \\
    &=& \Pr\left\{ \sup \left\{ n \geq 1 : \frac{\bar{X}_{ij}(n)-\mu_{ij}}{\sigma_{ij}} \leq  \frac{b_\delta-\mu_{ij}}{\sigma_{ij}}\right\} > r \right\}
    \leq  2 \exp\left( -r \frac{(\mu_{ij} - b_\delta)^2}{2\sigma_{ij}^2} \right).  \quad  \quad
\end{eqnarray}
The same bound is trivial when $r=0$. For scenarios with $\mu_{ij}\leq b_\delta$, we simply use $\Pr\{L_{ij}(b_\delta)>r\}\leq1$. Combining these bounds with Equation~\eqref{eq:productbound} yields
\begin{eqnarray}
\label{eq:part1}
    \Pr\left\{ \min_{j=1,\dots,m} L_{ij}(b_\delta) > r \right\} \leq \prod_{j: \mu_{ij} > b_\delta} \left[ 2 \exp\left( -r \frac{(\mu_{ij} - b_\delta)^2}{2\sigma_{ij}^2} \right) \right] 
= 2^{M_i} \exp\left( -r \sum_{j: \mu_{ij} > b_\delta} \frac{(\mu_{ij} - b_\delta)^2}{2\sigma_{ij}^2} \right),
\end{eqnarray}
where $M_i = \left| \left\{ j : \mu_{ij} > b_\delta \right\} \right|$ denotes the number of scenarios under alternative $i$ that exceed the boundary.

For the second term on the right-hand side of \eqref{eq: twoparts}, since $\mu_{1j} \leq \mu_{11} < b_\delta$ for all $j$, Lemma~\ref{lem: lastexit_tail} applies directly to each $U_{1j}(b_\delta)$, yielding:
\begin{eqnarray}
\label{eq:part2}
    \notag \Pr\left\{ U_{1j}(b_\delta) > r \right\}  & = &  \Pr\left\{ \sup \{ n \geq 1 : \bar{X}_{1j}(n) \geq b_\delta \} > r \right\}  \\
    &=& \Pr\left\{ \sup \left\{ n \geq 1 : \frac{\bar{X}_{1j}(n)-\mu_{1j}}{\sigma_{1j}} \geq  \frac{b_\delta-\mu_{1j}}{\sigma_{1j}}\right\} > r \right\}
    \leq  2 \exp\left( -r \frac{(b_\delta-\mu_{1j})^2}{2\sigma_{1j}^2} \right). \quad \quad 
\end{eqnarray}
For $r=0$, this bound is again trivial. Substituting Equations~\eqref{eq:part1} and \eqref{eq:part2} into Equation~\eqref{eq: twoparts} gives the stated PICS upper bound. Since $r$ grows linearly with $N$, the bound decays exponentially.
  \hfill \Halmos
  
\end{proof}

\subsection{Proof of Proposition~\ref{prop: allocation}}
\label{subsec: proof_prop_allocation}

\begin{proof}{Proof.}
Fix any \(b_\delta\in(\mu_{11},\mu_{21})\). Because
\(\mu_{1j}\leq\mu_{11}<b_\delta\), Lemma~\ref{lem: finite} implies that
\(U_{1j}(b_\delta)<\infty\) almost surely for every \(j\). Similarly,
because \(\mu_{i1}\geq\mu_{21}>b_\delta\), we have
\(L_{i1}(b_\delta)<\infty\) almost surely for every \(i\geq2\), and hence
\(\min_j L_{ij}(b_\delta)<\infty\). Since there are finitely many
alternatives and scenarios,
\begin{equation}
\label{eq:mstepnonbestfinite}
S(b_\delta)
=\sum_{i=2}^k\min_{j=1,\ldots,m}L_{ij}(b_\delta)
+\sum_{j=1}^mU_{1j}(b_\delta)
<\infty
\quad\text{almost surely}.
\end{equation}

By Lemma~\ref{lem: sufficient_time}, the total number of \(m\)-steps
allocated to the non-best alternatives is bounded by \(S(b_\delta)\).
Because exactly one \(m\)-step is performed in every round, after \(T\)
rounds,
\[
n_1^m(T)
=T-\sum_{i=2}^k n_i^m(T)
\geq T-S(b_\delta).
\]
Therefore, \(n_1^m(T)\to\infty\) almost surely as \(T\to\infty\). Every
scenario of alternative 1 receives one observation whenever alternative
1 receives an \(m\)-step. Consequently,
\(\lim_{N\to\infty}n_{1j}(N)=\infty\) almost surely for every
\(j=1,\ldots,m\), proving Part~(1).
Finally, in every round, each non-best alternative is sampled either in
the \(m\)-step, if it is the current best, or in the \(k\)-step otherwise.
Thus, its total sample size increases by at least one in every round, and
\(\lim_{N\to\infty}\sum_{j=1}^m n_{ij}(N)=\infty\) almost surely for every
\(i=2,\ldots,k\). This proves Part~(2). \hfill\Halmos
\end{proof}

\subsection{Proof of Theorem~\ref{thm: additive}}
\label{subsec: proof_thm_additive}

\subsubsection{Preliminaries}
\label{subsubsec:lemmas}

We first prepare the following lemmas.

\begin{lemma}[Attainment of the Minimum for a Convergent Sequence]
\label{lem:min_attainment}
Let \(\{a_n\}_{n=1}^\infty\) be a sequence of real numbers that converges
to a finite limit \(\mu\). Suppose that the sequence contains at least
one term strictly less than \(\mu\). Then, the minimum of the sequence is
attained at a finite index. That is, there exists a finite
\(K\in\mathbbm N^+\) such that
$
a_K=\min_{n\geq1}a_n.
$
\end{lemma}

\begin{proof}{Proof.}
Since \(\{a_n\}\) converges to \(\mu\), it is bounded. Let
\(M=\inf_{n\geq1}a_n\). By assumption, there exists at least one index
\(n\) such that \(a_n<\mu\), so \(M<\mu\). Choose
\(\epsilon=(\mu-M)/2>0\). By convergence, there exists a finite integer
\(N\) such that, for every \(n>N\),
$
|a_n-\mu|<\epsilon,
$
which implies
\(a_n>\mu-\epsilon=(M+\mu)/2>M\). Therefore, all terms in the tail
\(\{a_n\}_{n>N}\) are strictly greater than \(M\). It follows that the
infimum must be attained within the finite set
\(\{a_1,\ldots,a_N\}\). Hence, there exists
\(K\in\{1,\ldots,N\}\) such that
\(a_K=\min_{n\geq1}a_n\). \hfill\Halmos
\end{proof}

\begin{lemma}[Law of the Iterated Logarithm; \citealt{itemHartman1941}]
\label{thm:lil}
Let \(\{X_n\}_{n=1}^\infty\) be a sequence of i.i.d. random variables
with finite mean \(\E[X_n]=\mu\) and finite, positive variance
\(\mathrm{Var}[X_n]=\sigma^2\), and let
\(\bar X_n=n^{-1}\sum_{i=1}^nX_i\). Then, almost surely,
\[
\limsup_{n\to\infty}
\frac{\bar X_n-\mu}
{\sigma\sqrt{2\log\log n/n}}
=1
\quad\text{and}\quad
\liminf_{n\to\infty}
\frac{\bar X_n-\mu}
{\sigma\sqrt{2\log\log n/n}}
=-1.
\]
\end{lemma}

The following result is an immediate consequence.

\begin{lemma}[Infinite Visits Below the Mean]
\label{lem:below_mean}
Under the same conditions as Lemma~\ref{thm:lil}, the sample mean
\(\bar X_n\) is strictly less than \(\mu\) for infinitely many values of
\(n\) almost surely. Formally,
\[
\Pr\left\{
\left|\left\{n\in\mathbbm N^+:\bar X_n<\mu\right\}\right|
=\infty
\right\}=1.
\]
\end{lemma}

\begin{proof}{Proof.}
If \(\bar X_n<\mu\) occurred only finitely often, then
\(\bar X_n-\mu\geq0\) for all sufficiently large \(n\). This would
contradict the negative limit inferior in Lemma~\ref{thm:lil}.
\hfill\Halmos
\end{proof}

Combining Lemmas~\ref{lem:min_attainment} and~\ref{lem:below_mean}
gives the following tail-minimum property.

\begin{lemma}[Properties of the Tailed Sample Mean Process]
\label{lem:tailed_min}
Let \(\{X_n\}_{n=1}^\infty\) be a sequence of i.i.d. random variables
with finite mean \(\E[X_n]=\mu\) and finite, positive variance
\(\mathrm{Var}[X_n]=\sigma^2\), and let
\(\bar X_n=n^{-1}\sum_{i=1}^nX_i\). Then, almost surely, simultaneously
for every finite \(n'\in\mathbbm N^+\),

\begin{enumerate}
    \item the infimum of the tail sequence is strictly less than the
    mean, i.e.,
$
    \inf_{n\geq n'}\bar X_n<\mu;
$
    \item this infimum is attained at a finite index. That is, there
    exists a finite \(K(n')\geq n'\) such that
$
    \bar X_{K(n')}
    =\min_{n\geq n'}\bar X_n
    <\mu.
$
\end{enumerate}
\end{lemma}

\begin{proof}{Proof.}
We argue pathwise. Fix a sample path \(\omega\) such that
\(\bar X_n(\omega)\to\mu\) and
\(\bar X_n(\omega)<\mu\) for infinitely many \(n\). By the strong law
of large numbers and Lemma~\ref{lem:below_mean}, such sample paths occur
with probability one.

Fix any finite \(n'\in\mathbbm N^+\). Because
\(\bar X_n(\omega)<\mu\) infinitely often, the tail sequence
\(\{\bar X_n(\omega)\}_{n\geq n'}\) contains at least one term strictly
less than \(\mu\). Moreover, this tail sequence converges to \(\mu\).
Applying Lemma~\ref{lem:min_attainment} to the tail sequence shows that
its minimum is strictly less than \(\mu\) and is attained at a finite
index \(K(n')\). 
Because the set of possible starting indices \(n'\) is countable, the
result holds simultaneously for every finite \(n'\) almost surely.
\hfill\Halmos
\end{proof}

For completeness, we also establish that two independent normal
sample-mean processes almost surely never take exactly the same value.

\begin{lemma}[Almost Sure Inequality of Normal Sample Means]
\label{lem:inequality}
Let \(\{X_n\}_{n=1}^\infty\) and \(\{Y_m\}_{m=1}^\infty\) be two
mutually independent i.i.d. sequences, where
\(X_n\sim\mathcal N(\mu_X,\sigma_X^2)\) and
\(Y_m\sim\mathcal N(\mu_Y,\sigma_Y^2)\), with
\(\sigma_X^2>0\) and \(\sigma_Y^2>0\). Let
$
\bar X_n=\frac{1}{n}\sum_{i=1}^nX_i$ and $
\bar Y_m=\frac{1}{m}\sum_{j=1}^mY_j.
$
Then,
$
\Pr\left\{
\bigcup_{n=1}^\infty\bigcup_{m=1}^\infty
\left\{\bar X_n=\bar Y_m\right\}
\right\}=0.
$
\end{lemma}


\subsubsection{Proof of Theorem~\ref{thm: additive}}
\label{subsubsec:proof}

\begin{proof}{Proof.}
We first collect the probability-one properties used in the proof. Let
\(\Omega\) be the event on which:

\begin{enumerate}
    \item the conclusions of Proposition~\ref{prop: allocation} hold;
    \item every non-best alternative receives only finitely many
    \(m\)-steps, as established by Lemma~\ref{lem: sufficient_time} and
    Equation~\eqref{eq:mstepnonbestfinite};
    \item the conclusions of Lemma~\ref{lem:tailed_min} hold for every
    scenario and every finite starting index; and
    \item for every alternative \(i\), every pair \(j\neq\ell\), and all
    positive integers \(n\) and \(r\),
    \(\bar X_{ij}(n)\neq\bar X_{i\ell}(r)\).
\end{enumerate}

The first three properties hold almost surely by the corresponding
results. The fourth follows from Lemma~\ref{lem:inequality}, together
with a finite union over all alternatives and scenario pairs. Hence,
\(\Pr\{\Omega\}=1\). We fix an arbitrary sample path in \(\Omega\) for
the remainder of the proof.

By Proposition~\ref{prop: allocation}, all \(m\) scenarios of the best
alternative are sampled infinitely often:
\begin{equation}
\label{eq:m_infinite}
\sum_{j=1}^m
\mathbbm{1}_{\lim_{N\to\infty}n_{1j}(N)=\infty}
=m.
\end{equation}
Moreover, the total sample size of every non-best alternative tends to
infinity. Because each alternative has only finitely many scenarios, at
least one scenario of every non-best alternative must therefore be
sampled infinitely often. It remains to show that no two scenarios of a
non-best alternative can both be sampled infinitely often.

Fix a non-best alternative \(i\). Because it receives only finitely many
\(m\)-steps, there exists a finite round \(T_i\) after which alternative
\(i\) is sampled only through the \(k\)-step. From that point onward,
the procedure greedily samples the scenario of alternative \(i\) having
the largest current sample mean.
Consider two distinct scenarios \(j\) and \(\ell\) of alternative \(i\).
Define their tail minima after round \(T_i\) as
\[
M_{ij}
=\min_{n\geq n_{ij}(T_i)}\bar X_{ij}(n),
\qquad
M_{i\ell}
=\min_{n\geq n_{i\ell}(T_i)}\bar X_{i\ell}(n).
\]
By Lemma~\ref{lem:tailed_min}, these minima are attained at finite
sample sizes. Thus, there exist finite indices
\(K_{ij}\geq n_{ij}(T_i)\) and
\(K_{i\ell}\geq n_{i\ell}(T_i)\) such that
\[
M_{ij}=\bar X_{ij}(K_{ij}),
\qquad
M_{i\ell}=\bar X_{i\ell}(K_{i\ell}).
\]
Moreover, by the fourth defining property of \(\Omega\),
\begin{equation}
\label{eq:minrelationdecouple}
M_{ij}\neq M_{i\ell}.
\end{equation}

Suppose, toward a contradiction, that scenarios \(j\) and \(\ell\) are
both sampled infinitely often. Without loss of generality, suppose that
\(M_{ij}<M_{i\ell}\). Because scenario \(j\) is sampled infinitely
often and each \(k\)-step allocation increases its sample size by one,
its sample size must eventually reach the finite index \(K_{ij}\). At
that point, its current sample mean equals \(M_{ij}\).
On the other hand, by the definition of \(M_{i\ell}\), the current and
all subsequent sample means of scenario \(\ell\) satisfy
\[
\bar X_{i\ell}(n)\geq M_{i\ell}>M_{ij}
=\bar X_{ij}(K_{ij}),
\qquad n\geq n_{i\ell}(T_i).
\]
Thus, once scenario \(j\) reaches \(K_{ij}\), scenario \(\ell\) has a
strictly larger sample mean in every subsequent round. Scenario \(j\)
can therefore never again be the empirical worst-case scenario of
alternative \(i\) and cannot be selected by the greedy \(k\)-step.
Because alternative \(i\) receives no further \(m\)-steps, scenario
\(j\) is sampled only finitely often, contradicting the assumption. The
case \(M_{ij}>M_{i\ell}\) follows symmetrically.

Hence, for every non-best alternative \(i\), at most one scenario is
sampled infinitely often:
\begin{equation}
\label{eq:notwoinfinite}
\sum_{j=1}^m
\mathbbm{1}_{\lim_{N\to\infty}n_{ij}(N)=\infty}
\leq1,
\qquad i=2,\ldots,k.
\end{equation}
Proposition~\ref{prop: allocation} guarantees that at least one scenario
of each non-best alternative is sampled infinitely often. Therefore,
the inequality in Equation~\eqref{eq:notwoinfinite} is an equality.
Combining this result with Equation~\eqref{eq:m_infinite} gives
\[
\sum_{i=1}^k\sum_{j=1}^m
\mathbbm{1}_{\lim_{N\to\infty}n_{ij}(N)=\infty}
=m+(k-1)=k+m-1.
\]
Because the sample path was chosen arbitrarily from the probability-one
event \(\Omega\), the result holds almost surely. \hfill\Halmos
\end{proof}

\subsection{Proof of Proposition \ref{prop: possible_limiting} and Theorem \ref{thm: non_necessity}}
\label{subsec: proof_thm_non_necessity}

To prove the result, we first prepare the following lemma.
\begin{lemma}
\label{lem:zero_exit_times}
For the last exit times $U(b)$ and $L(-b)$ defined in Equation \eqref{eq: general_def}, we have that for any boundary $b > 0$,
$$
\Pr\left\{ U(b) = 0 \right\}
= \Pr\left\{ L(-b)=0 \right\}
\geq 1 - \exp\left( -\frac{b^2}{2} \right).
$$
\end{lemma}

\begin{proof}{Proof.} First, notice that by the definition of $U(b)$ and $L(-b)$, 
$
\Pr\left\{ U(b) = 0 \right\} = \Pr\{\sup \{ n \geq 1 : \bar Z(n) \geq b \}=0\} =  \Pr\left\{ \bar{Z}(n) < b, \forall n \geq 1 \right\}
= \Pr\left\{ \bar{Z}(n) > -b, \forall n \geq 1 \right\} = \Pr\{\sup \{ n \geq 1 : \bar Z(n) < -b \}=0\} = \Pr\left\{ L(-b) = 0 \right\}$, due to the symmetry of the standard normal distribution around zero.  Then, we have
    \begin{eqnarray*}
  \Pr\left\{ \bar Z(n) > -b, \forall n \geq 1 \right\} 
 &=& \exp\left(-\sum_{n=1}^\infty \frac{1}{n} \left(1 - \Phi(\sqrt{n}b)\right)\right) \geq  \exp\left(-\sum_{n=1}^\infty \frac{1}{n} \exp\left(-\frac{n b^2}{2} \right) \right)  = 1 - \exp\left(-\frac{b^2}{2} \right),
\end{eqnarray*}
where the first equality follows from the result in Lemma 2 of \cite{itemLi2024},  the inequality uses the well-known Gaussian tail bound $1 - \Phi(x) \leq \exp(-x^2/2)$ for all $x > 0$,  and the final identity applies the expansion $\sum_{n=1}^\infty   e^{-nx}/n = -\ln(1 - e^{-x})$ for $x > 0$. This result of interest is proved. \hfill \Halmos
\end{proof}

\begin{proof}{\textbf{Proof of Proposition \ref{prop: possible_limiting} and Theorem \ref{thm: non_necessity}.}} The result can be proved surprisingly directly using a boundary-crossing perspective. Fix any non-best alternative $i=2,\dots,k$ and any scenario $j$ satisfying $\mu_{ij}>\mu_{11}$. Choose a boundary value $b_\delta\in(\mu_{11},\mu_{ij})$. Suppose that the sample means $\bar X_{1\ell}(n)$ of \textit{every} scenario $(1,\ell)$ of the best alternative remain below $b_\delta$ for all $n\geq1$ (i.e., $U_{1\ell}(b_\delta)=0$), while the sample mean $\bar X_{ij}(n)$ remains above $b_\delta$ for all $n\geq1$ (i.e., $L_{ij}(b_\delta)=0$). Suppose also that the initial observation of every other scenario $(i,\ell)$, $\ell\neq j$, lies below $b_\delta$.

Under these events, the empirical worst-case performance of alternative 1 always lies below $b_\delta$, while that of alternative $i$ always lies above $b_\delta$. Hence, alternative $i$ is never selected for an $m$-step. Moreover, scenario $(i,j)$ always dominates the other scenarios of alternative $i$ in the greedy $k$-step and is therefore sampled in every round. Consequently, $n_{ij}(t)\rightarrow\infty$, whereas $n_{i\ell}(t)=1$ for all $\ell\neq j$ and all $t\geq1$. Formalizing this insight, define
\begin{eqnarray*}
\mathcal E_{ij}(b_\delta)
=
\left\{L_{ij}(b_\delta)=0\right\}
\cap\bigcap_{\ell\neq j}\left\{\bar X_{i\ell}(1)\leq b_\delta\right\}
\cap\bigcap_{\ell=1}^m\left\{U_{1\ell}(b_\delta)=0\right\}.
\end{eqnarray*}
Then,
\begin{eqnarray}
\label{eq:formal_insights}
\notag  \Pr\left\{\lim_{N\rightarrow\infty}n_{ij}(N)=\infty,\
\lim_{N\rightarrow\infty}n_{i\ell}(N)<\infty,\ \forall\,\ell\neq j\right\} &\geq&\Pr\left\{\mathcal E_{ij}(b_\delta)\right\}\\
&=&\Pr\left\{L_{ij}(b_\delta)=0\right\}
\prod_{\ell\neq j}\Pr\left\{\bar X_{i\ell}(1)\leq b_\delta\right\}
\prod_{\ell=1}^m\Pr\left\{U_{1\ell}(b_\delta)=0\right\},
\end{eqnarray}
where the equality follows from the independence of the sampling processes across scenarios.

From Lemma \ref{lem:zero_exit_times}, we know that for every $\ell=1,\dots,m$, since $\mu_{1\ell}\leq\mu_{11}<b_\delta$, we have
\begin{eqnarray} 
    \label{eq:boundU}
    \Pr\left\{ U_{1\ell}(b_\delta) = 0 \right\}= \Pr\left\{ \sup \left\{ n \geq 1 : \frac{\bar{X}_{1\ell}(n)-\mu_{1\ell}}{\sigma_{1\ell}} \geq  \frac{b_\delta-\mu_{1\ell}}{\sigma_{1\ell}}\right\} = 0 \right\}  \geq 1- \exp\left(-\frac{( b_\delta-\mu_{1\ell})^2}{2 \sigma_{1\ell}^2} \right) := b_{1\ell}.\quad \quad \quad 
\end{eqnarray}
Similarly, because $\mu_{ij}>b_\delta$,
\begin{eqnarray} 
    \label{eq:boundL}
    \Pr\left\{ L_{ij}(b_\delta) = 0 \right\}= \Pr\left\{ \sup \left\{ n \geq 1 : \frac{\bar{X}_{ij}(n)-\mu_{ij}}{\sigma_{ij}} \leq  \frac{b_\delta-\mu_{ij}}{\sigma_{ij}}\right\} = 0 \right\} \geq  \left(  1- \exp\left(-\frac{( b_\delta-\mu_{ij})^2}{2 \sigma_{ij}^2} \right) \right):= b_{ij}. \quad\quad \quad
\end{eqnarray}
Moreover, for every $\ell\neq j$,
$
\Pr\left\{\bar X_{i\ell}(1)\leq b_\delta\right\}
=\Phi\left(\frac{b_\delta-\mu_{i\ell}}{\sigma_{i\ell}}\right):=a_{i\ell}>0.
$
Substituting these bounds into Equation~\eqref{eq:formal_insights} gives
\begin{eqnarray*}
 \Pr\left\{\lim_{N\rightarrow\infty}n_{ij}(N)=\infty,\
\lim_{N\rightarrow\infty}n_{i\ell}(N)<\infty,\ \forall\,\ell\neq j\right\}  \geq b_{ij}\prod_{\ell\neq j}a_{i\ell}\prod_{\ell=1}^m b_{1\ell}>0,
\end{eqnarray*}
which proves the proposition. If $\mu_{ij}<\mu_{i1}$, then $j$ is not a true worst-case scenario. Every scenario $\ell$ satisfying $\mu_{i\ell}=\mu_{i1}$ therefore has $\ell\neq j$ and is sampled only finitely often on the event above. Because the event in Proposition~\ref{prop: possible_limiting} is contained in the event in Theorem~\ref{thm: non_necessity}, the same lower bound applies. This proves the theorem.
\hfill \Halmos
\end{proof}

\section{Technical Supplement to Section \ref{sec: general}}
\label{sec:analysis_GAA}

\subsection{Budget Allocation Analysis for GAA Procedures}
The analysis of the AA procedure heavily relies on the boundary-crossing analysis presented in Section~\ref{subsec:boundary}, particularly Lemma~\ref{lem: sufficient_time}. To prove analogous properties of the AA procedure under the GAA framework, we first establish a result similar to Lemma~\ref{lem: sufficient_time}. Specifically, we prove the following lemma.
\begin{lemma}
\label{lem: sufficient_time2} 
For GAA procedures satisfying Assumptions \ref{assu:exploration_m} and \ref{assu:exploration_k}, the following hold almost surely:
\begin{itemize}
    \item [(1)] for alternative 1, its number of $m$-steps allocated $r_1^m(t)$ satisfies
$
    \lim_{t \rightarrow \infty} r_1^m(t) = \infty.
$
\item [(2)] for each non-best alternative $i=2, \dots, k$, the number of $m$-steps allocated $r_i^m(t)$ satisfies 
$
     \lim_{t \rightarrow \infty} r_i^m(t) < \infty.
$
\end{itemize}
\end{lemma}

Ideally, we would like to establish the lemma using arguments similar to those in Section~\ref{subsec:boundary}. However, due to the adaptive $m$-step and $k$-step allocations in the GAA procedures, such a direct approach is no longer feasible. Instead, we must adopt a more involved proof. Before proceeding, we first summarize the various counters used in the analysis in Table~\ref{tab:notation_sets} below.

\begin{table}[htbp] 
    \small
    \centering
    \caption{Counters for the Sampling Process of GAA Procedures}
    \label{tab:notation_sets}
    \begin{tabular}{p{1.5cm}|p{6.5cm}|p{7.5cm}}
        \hline
        
        \hline
        
        \text{Notation} & \text{Definition} & \text{ Meaning} \\
        \hline

        \hline

        $r_i^m(t)$  & $\sum_{\tau=1}^t \mathbbm{1}_{  i = \hat{b}(\tau)  }$ & Number of rounds in which alternative $i$ is selected for the $m$-step (i.e.$i = \hat{b}(t)$) \\  
        \hline
        $r_i^k(t)$ & $\sum_{\tau=1}^t \mathbbm{1}_{  i \in K(\tau)  }$ & Number of rounds in which alternative $i$ is selected for the $k$-step (i.e.$i \in K(t)$)  \\  
        \hline
                $n_{ij}^m(t)$ &$\sum_{\tau=1}^t \Delta^m_{ij}(\tau)$ & Cumulative number of observations allocated to scenario $(i,j)$ from $m$-steps  \\
        \hline
        $n_{ij}^k(t)$ &$\sum_{\tau=1}^t \Delta^k_{ij}(\tau)$ & Cumulative number of observations allocated to scenario $(i,j)$ from $k$-steps  \\
        \hline
                        $n_{ij}(t)$ &$n_{ij}^m(t)+n_{ij}^k(t)+n_0$ & Cumulative number of observations allocated to scenario $(i,j)$ \\
        \hline
                        $n_{ij}^{k+}(t)$ & $\sum_{\tau=1}^t \mathbbm{1}_{  i \in K(\tau)  }\mathbbm{1}_{  \bar X_{ij}(n_{ij}(\tau-1)) \geq b_\delta }  \Delta^k_{ij} (\tau)$ & Cumulative number of observations allocated to $(i,j)$ in $k$-steps when its sample mean exceeds $b_\delta$ \\ 
        \hline
                                $n_{i}^{k+}(t)$ & $\sum_{\tau=1}^t \mathbbm{1}_{  i \in K(\tau)  } \mathbbm{1}_{  \max_{j=1,\dots,m} \bar X_{ij}(n_{ij}(\tau-1)) \geq b_\delta } \sum_{j=1}^m \Delta^k_{ij}(\tau)$ & Total number of observations allocated to alternative $i$ in $k$-steps when its current worst-case scenario has sample mean exceeding $b_\delta$ \\        
        \hline

        \hline
        
    \end{tabular} 
\end{table}

In Table~\ref{tab:notation_sets}, the last two counters involve conditional $k$-step allocations and will play a key role in the analysis. We first present a few useful results to support later proofs. Fix a boundary $b_\delta \in (\mu_{11}, \mu_{21})$. Recall that for each alternative $i = 1, 2, \dots, k$, in any round $t$, if $i$ is selected for the $k$-step, only its current empirical worst-case scenario is included in the budget allocation of $\Delta_k$, together with the worst-case scenarios of other $k$-step alternatives. Therefore, for any scenario $(i, j)$, if $\Delta^k_{ij}(\tau) > 0$, then $(i, j)$ must be the empirical worst-case scenario of alternative $i$ at round $\tau$, and $\Delta^k_{ij\prime}(\tau) = 0$ for all $j\prime \neq j$. Therefore, we have 
\begin{eqnarray*}
    n_{ij}^{k+}(t) = \sum_{\tau=1}^t \mathbbm{1}_{  i \in K(\tau)  }\mathbbm{1}_{  \bar X_{ij}(n_{ij}(\tau-1)) \geq b_\delta }  \Delta^k_{ij} (\tau) = \sum_{\tau=1}^t \mathbbm{1}_{  i \in K(\tau)  }\mathbbm{1}_{  \bar X_{ij}(n_{ij}(\tau-1)) \geq b_\delta }  \mathbbm{1}_{  \bar X_{ij}(n_{ij}(\tau-1)) = \max_{j=1,\dots,m} \bar X_{ij}(n_{ij}(\tau-1)) } \Delta^k_{ij} (\tau).  \quad  \quad  
\end{eqnarray*} 
Let $j^*_i(\tau) = \argmax_{j=1,\dots,m} \bar X_{ij}(n_{ij}(\tau-1))$. It follows that
\begin{eqnarray}
\label{eq:kplusdecouple}
    \sum_{j=1}^m n_{ij}^{k+}(t) = \sum_{\tau=1}^t \mathbbm{1}_{  i \in K(\tau)  } \mathbbm{1}_{  \bar X_{ij^*_i(\tau)}(n_{ij^*_i(\tau)}(\tau-1)) \geq b_\delta } \Delta^k_{ij^*_i(\tau)}(\tau) = \sum_{\tau=1}^t \mathbbm{1}_{  i \in K(\tau)  } \mathbbm{1}_{  \max_{j=1,\dots,m} \bar X_{ij}(n_{ij}(\tau-1)) \geq b_\delta } \sum_{j=1}^m \Delta^k_{ij}(\tau) =  n_{i}^{k+}(t). \quad \quad \quad   
\end{eqnarray} 
Furthermore, we note another useful observation for the best alternative. For each scenario $(1,j)$ of alternative $1$, we have the following almost sure uniform bound 
\begin{eqnarray}
\label{eq:controlofkplus}
    n_{1j}^{k+}(t) \leq U_{1j}(b_\delta),\quad \forall t.
\end{eqnarray}
To see this, consider any $k$-step observation counted by $n_{1j}^{k+}(t)$. At the beginning of that round, its pre-allocation sample size $s=n_{1j}(\tau-1)$ satisfies $\bar X_{1j}(s)\geq b_\delta$. By the definition of $U_{1j}(b_\delta)$, this implies $s\leq U_{1j}(b_\delta)$. Moreover, because $\Delta_{1j}^k(\tau)\in\{0,1\}$ under Assumption~\ref{assu:exploration_k}, distinct counted allocations correspond to distinct pre-allocation sample sizes. Therefore, at most $U_{1j}(b_\delta)$ such observations can be counted, proving Equation~\eqref{eq:controlofkplus}. We now proceed to the proof of the lemma.

\vspace{10pt}
\begin{proof}{Proof of \textbf{Lemma \ref{lem: sufficient_time2}}.}
Since each round of the GAA procedure includes exactly one $m$-step selection, it holds that
$
\Pr\left\{ \exists i \in \{1, \dots, k\}: \lim_{t \to \infty} r_i^m(t) = \infty \right\} = 1.
$
Then, using the union bound, we have
\begin{eqnarray}
\label{eq:unionbound}
    \notag \Pr\left\{ \lim_{t \to \infty} r_1^m(t) = \infty,\forall i \geq 2: \lim_{t \to \infty} r_i^m(t) < \infty \right\} & =& \Pr\left\{ \exists i \in \{1, \dots, k\}: \lim_{t \to \infty} r_i^m(t) = \infty \right\} - \Pr\left\{ \exists i \geq 2: \lim_{t \to \infty} r_i^m(t) = \infty \right\} \\
    &\geq &  1 - \sum_{i=2}^k \Pr\left\{ \lim_{t \to \infty} r_i^m(t) = \infty \right\}.
\end{eqnarray}
It therefore suffices to show that each non-best alternative receives only finitely many $m$-steps almost surely.
    
    For notational convenience, let
    $\widehat\mu_i(t)=\max_{j=1,\dots,m}\bar X_{i j}(n_{i j}(t))$ for each alternative $i = 1, 2, \dots, k$, and define 
    $$A_i=\left\{\lim_{t\to\infty}r_i^m(t)=\infty\right\}, \forall i \neq 1, \text{  and  } B=\left\{\lim_{t\to\infty}r_1^m(t)=\infty\right\}.$$ Also, define 
    $$G_i=\{\exists T<\infty:\widehat\mu_i(t)>b_\delta,\ \forall t\geq T\}, \forall i \neq 1, \text{  and  } G_1=\{\exists T<\infty:\widehat\mu_1(t)<b_\delta,\ \forall t\geq T\}.$$

    Under Assumption~\ref{assu:exploration_m}, if $r_i^m(t)\to\infty$, then every scenario of alternative $i$ is sampled infinitely often through the $m$-steps. In particular,
    $
    \Pr\{A_i\}
    =
    \Pr\left\{
    A_i,\ \lim_{t\to\infty}n_{i1}^m(t)=\infty
    \right\}.
    $
    Since $\mu_{i1}\geq\mu_{21}>b_\delta$, Lemma~\ref{lem: finite} implies that $L_{i1}(b_\delta)<\infty$ almost surely. Therefore,
    \begin{eqnarray*}
    \Pr\{A_i\}
    =
    \Pr\left\{
    A_i,\ \lim_{t\to\infty}n_{i1}^m(t)=\infty,\
    L_{i1}(b_\delta)<\infty
    \right\}.
    \end{eqnarray*}
    From the definition of $L_{i1}(b_\delta)$, it follows that
    \begin{eqnarray*}
    \label{eq:step1}
    \notag
    \Pr\{A_i\}
    \leq
    \Pr\left\{
    A_i,\ \exists T<\infty:
    n_{i1}^m(t)>L_{i1}(b_\delta),\ \forall t\geq T
    \right\} 
    &\leq&
    \Pr\left\{
    A_i,\ \exists T<\infty:
    n_{i1}(t)>L_{i1}(b_\delta),\ \forall t\geq T
    \right\}\\
    \notag
    &\leq&
    \Pr\left\{
    A_i,\ \exists T<\infty:
    \bar X_{i1}(n_{i1}(t))>b_\delta,\ \forall t\geq T
    \right\}\\
    &\leq&
    \Pr\left\{
    A_i,\ \exists T<\infty:
    \widehat\mu_i(t)>b_\delta,\ \forall t\geq T
    \right\}
    =
    \Pr\{A_i\cap G_i\},
    \end{eqnarray*}
    where the second inequality follows from $n_{i1}(t)\geq n_{i1}^m(t)$. Since $A_i\cap G_i\subseteq A_i$, it implies $\Pr\{A_i\}
    =
    \Pr\{A_i\cap G_i\}$, and hence, 
    \begin{eqnarray}
    \label{eq:nonbest_eventually_above}
    \Pr\{A_i\cap G_i^c\}=0.
    \end{eqnarray}
    
    We next derive a corresponding observation for the best alternative. Under Assumption~\ref{assu:exploration_m},
    $
    \Pr\{B\}
    =
    \Pr\left\{
    B,\ \lim_{t\to\infty}n_{1j}^m(t)=\infty,\ \forall j
    \right\}.
    $
   Since $\mu_{1j}\leq\mu_{11}<b_\delta$, Lemma~\ref{lem: finite} implies that $U_{1j}(b_\delta)<\infty$ almost surely for every $j$. Thus,
    \begin{eqnarray*}
    \Pr\{B\}
    =
    \Pr\left\{
    B,\ \lim_{t\to\infty}n_{1j}^m(t)=\infty,\
    U_{1j}(b_\delta)<\infty,\ \forall j
    \right\}.
    \end{eqnarray*}
    It follows from the definition of $U_{1j}(b_\delta)$ that
    \begin{eqnarray*}
    \label{eq:best_eventually_below_m}
    \notag
    \Pr\{B\}
    \leq
    \Pr\left\{
    B,\ \exists T<\infty:
    n_{1j}^m(t)>U_{1j}(b_\delta),\
    \forall j,\ \forall t\geq T
    \right\} 
    &\leq&
    \Pr\left\{
    B,\ \exists T<\infty:
    n_{1j}(t)>U_{1j}(b_\delta),\
    \forall j,\ \forall t\geq T
    \right\}\\
    &\leq&
    \Pr\left\{
    B,\ \exists T<\infty:
    \widehat\mu_1(t)<b_\delta,\ \forall t\geq T
    \right\}
    =
    \Pr\{B\cap G_1\}.
    \end{eqnarray*}
    Since $B\cap G_1\subseteq B$, we conclude that $\Pr\{B\}
    =
    \Pr\{B\cap G_1\}$, and hence, 
    \begin{eqnarray}
    \label{eq:best_eventually_below}
    \Pr\{B\cap G_1^c\}=0.
    \end{eqnarray}
    
    We now decompose $A_i$ according to whether alternative 1 receives infinitely many $m$-steps:
    \begin{eqnarray}
    \label{eq:two_case_decomposition}
    \Pr\{A_i\}
    =
    \Pr\{A_i\cap B\}
    +
    \Pr\{A_i\cap B^c\}.
    \end{eqnarray}
    For the first case, Equations~\eqref{eq:nonbest_eventually_above} and \eqref{eq:best_eventually_below} give
    \begin{eqnarray}
        \label{eq:first_case_zero}
    \Pr\{A_i\cap B\}
    =
    \Pr\{A_i\cap B\cap G_i\cap G_1\}
    \leq
    \Pr\{A_i\cap G_i\cap G_1\}
    =0.
    \end{eqnarray}
    Indeed, on $G_i\cap G_1$, for all sufficiently large $t$,
    $
    \widehat\mu_1(t)<b_\delta<\widehat\mu_i(t).
    $
    Alternative $i$ therefore cannot be identified as the current best in any sufficiently late round, contradicting $A_i$. Hence, $ \Pr\{A_i\cap B\}=0$.
    
    We next consider $A_i\cap B^c$. Whenever alternative $i$ is selected for an $m$-step, alternative 1 is included in the $k$-step. Therefore,
    $
    A_i
    \subseteq
    \left\{
    \lim_{t\to\infty}r_1^k(t)=\infty
    \right\}.
    $
    By Assumption~\ref{assu:exploration_k},
    \begin{eqnarray}
    \label{eq:best_k_infinite}
    \Pr\{A_i\cap B^c\}
    =
    \Pr\left\{
    A_i\cap B^c,\
    \lim_{t\to\infty}n_1^k(t)=\infty
    \right\}.
    \end{eqnarray}
    
    Meanwhile, Equations~\eqref{eq:kplusdecouple} and \eqref{eq:controlofkplus} imply that, almost surely,
    $
    n_1^{k+}(t)
    =
    \sum_{j=1}^m n_{1j}^{k+}(t)
    \leq
    \sum_{j=1}^m U_{1j}(b_\delta)
    <\infty, \forall t,
        $
    where the last inequality follows from Lemma~\ref{lem: finite}. Consequently,
    \begin{eqnarray}
    \label{eq:kplus_probability_zero}
    \Pr\left\{
    \lim_{t\to\infty}n_1^{k+}(t)=\infty
    \right\}=0.
    \end{eqnarray}
    
    We further have
    \begin{eqnarray}
    \label{eq:kplus_implication}
    B^c
    \cap
    \left\{\lim_{t\to\infty}n_1^k(t)=\infty\right\}
    \cap G_1^c
    \subseteq
    \left\{\lim_{t\to\infty}n_1^{k+}(t)=\infty\right\}.
    \end{eqnarray}
        To verify this inclusion, suppose that $B^c$ and $n_1^k(t)\to\infty$ hold, but $n_1^{k+}(t)<\infty$. After the last $m$-step allocated to alternative 1 and the last observation counted by $n_1^{k+}$, every subsequent $k$-step observation allocated to alternative 1 must occur when its pre-allocation worst-case sample mean is below $b_\delta$. If any such observation causes its worst-case sample mean to become at least $b_\delta$, that sample mean remains unchanged until the next $k$-step observation allocated to alternative 1. Such a subsequent observation must exist because $n_1^k(t)\to\infty$, and it would be counted by $n_1^{k+}$, a contradiction. Hence, $\widehat\mu_1(t)<b_\delta$ eventually, so $G_1$ occurs. Taking the contrapositive proves Equation~\eqref{eq:kplus_implication}.
    Combining Equations~\eqref{eq:kplus_probability_zero} and \eqref{eq:kplus_implication} yields
    \begin{eqnarray}
    \label{eq:best_eventually_below_k}
    \Pr\left\{
    B^c,\
    \lim_{t\to\infty}n_1^k(t)=\infty,\
    G_1^c
    \right\}=0.
    \end{eqnarray}

    Using Equations~\eqref{eq:best_k_infinite} and \eqref{eq:best_eventually_below_k}, we obtain
    $
    \Pr\{A_i\cap B^c\}
    =
    \Pr\left\{
    A_i\cap B^c,\
    \lim_{t\to\infty}n_1^k(t)=\infty,\
    G_1
    \right\}.
    $
    Equation~\eqref{eq:nonbest_eventually_above} then gives
    \begin{eqnarray}
        \label{eq:second_case_zero}
    \Pr\{A_i\cap B^c\}
    =
    \Pr\left\{
    A_i\cap B^c,\
    \lim_{t\to\infty}n_1^k(t)=\infty,\
    G_1\cap G_i
    \right\}
    \leq
    \Pr\{A_i\cap G_1\cap G_i\}
    =0.
    \end{eqnarray}
    Combining Equations~\eqref{eq:two_case_decomposition}, \eqref{eq:first_case_zero}, and \eqref{eq:second_case_zero}, we conclude that
    \begin{eqnarray*}
    \Pr\left\{
    \lim_{t\to\infty}r_i^m(t)=\infty
    \right\}
    =
    \Pr\{A_i\}
    =0,\qquad i=2,\dots,k.
    \end{eqnarray*}
    Substituting this result into Equation~\eqref{eq:unionbound} proves both statements of the lemma.
    \hfill\Halmos
    \end{proof}
\vspace{4pt}

A direct consequence of Lemma~\ref{lem: sufficient_time2} is the following result, which extends Proposition~\ref{prop: allocation} from AA to GAA.
\begin{proposition}
\label{prop: allocation2}
For GAA procedures satisfying Assumptions \ref{assu:exploration_m} and \ref{assu:exploration_k}, it holds that
    \begin{itemize}
        \item[(1)] For $j=1, \dots, m$, $\lim_{N\rightarrow \infty} n_{1j} = \infty$ almost surely. 
        \item[(2)] For alternative $i=2, \dots, k$, $\lim_{N\rightarrow \infty} \sum_{j=1}^m n_{ij} = \infty$  almost surely. 
    \end{itemize}
\end{proposition}
\begin{proof}{Proof.}
    Part (1) follows from Part (1) of Lemma~\ref{lem: sufficient_time2} and Assumption~\ref{assu:exploration_m}. For Part (2), Lemma~\ref{lem: sufficient_time2} implies that for each non-best alternative $i = 2, \dots, k$, we have $\lim_{t \to \infty} r_i^k(t) = \infty$.  Then, the result follows from Assumption \ref{assu:exploration_k}. \hfill\Halmos
\end{proof}

\subsection{Proof of Theorem \ref{thm: properties_general}}
\label{subsec: proof_GAA_additive}
\begin{proof}{Proof.}
We first establish consistency. Let
\[
\mathcal C_t
=\left\{r_1^m(t)>r_i^m(t),\ \forall i=2,\dots,k\right\}.
\]
On \(\mathcal C_t\), GAA selects alternative 1. Lemma~\ref{lem: sufficient_time2} implies that, almost surely, \(r_1^m(t)\to\infty\), whereas \(r_i^m(t)\) converges to a finite limit for every \(i\geq2\). Hence, on this probability-one event, there exists a finite random round after which \(\mathcal C_t\) always occurs. Therefore, \(\mathbbm{1}_{\mathcal C_t}\to1\) almost surely. By the dominated convergence theorem,
\[
\lim_{t\to\infty}\Pr\{\mathcal C_t\}=1.
\]
Because the PCS at round \(t\) is at least \(\Pr\{\mathcal C_t\}\), and the number of completed rounds tends to infinity as \(N\to\infty\), the GAA procedure is consistent.

We next prove additivity. Consider the probability-one event on which the conclusions of Lemma~\ref{lem: sufficient_time2}, Proposition~\ref{prop: allocation2}, Lemma~\ref{lem:tailed_min}, and Lemma~\ref{lem:inequality} all hold simultaneously. We argue pathwise on this event. Proposition~\ref{prop: allocation2} first ensures that all \(m\) scenarios of the best alternative are sampled infinitely often.
Fix a non-best alternative \(i\). Lemma~\ref{lem: sufficient_time2} implies that there exists a finite round \(T_i\) after which \(i\) receives no further \(m\)-steps. Thereafter, only its current empirical worst-case scenario can receive observations through the \(k\)-step. Proposition~\ref{prop: allocation2} also ensures that the total sample size of alternative \(i\) diverges, so at least one of its scenarios is sampled infinitely often.

It remains to show that two scenarios cannot both be sampled infinitely often. Fix two distinct scenarios \(j\) and \(\ell\), and define
\[
M_{ij}=\min_{n\geq n_{ij}(T_i)}\bar X_{ij}(n),
\qquad
M_{i\ell}=\min_{n\geq n_{i\ell}(T_i)}\bar X_{i\ell}(n).
\]
By Lemma~\ref{lem:tailed_min}, these minima are attained at finite sample sizes \(K_{ij}\) and \(K_{i\ell}\). By Lemma~\ref{lem:inequality}, \(M_{ij}\neq M_{i\ell}\). Without loss of generality, suppose that \(M_{ij}<M_{i\ell}\).
If scenario \(j\) were sampled infinitely often, the unit-allocation condition \(\Delta_{ij}^k(t)\in\{0,1\}\) would imply that its sample size eventually reaches \(K_{ij}\). At that point, its sample mean equals \(M_{ij}\), while the current and all subsequent sample means of scenario \(\ell\) are at least \(M_{i\ell}>M_{ij}\). Scenario \(j\) can therefore never again be the empirical worst-case scenario and cannot receive another \(k\)-step observation. Since alternative \(i\) receives no further \(m\)-steps, scenario \(j\) must then be sampled only finitely often, a contradiction. The case \(M_{ij}>M_{i\ell}\) is symmetric.

Thus, at most one scenario of each non-best alternative is sampled infinitely often. Since Proposition~\ref{prop: allocation2} guarantees that at least one is, exactly one scenario of each non-best alternative is sampled infinitely often. Together with the \(m\) scenarios of the best, the total number of scenarios sampled infinitely often is \(k+m-1\) almost surely. \hfill\Halmos
\end{proof}

\subsection{Proof of Proposition \ref{prop: possible_limiting_general} and Theorem \ref{thm: non_necessity_general}}
\label{subsec: proof_thm_non_necessity_general}
\begin{proof}{Proof.}
Fix any non-best alternative \(i\), scenario \(j\) satisfying \(\mu_{ij}>\mu_{11}\), and boundary \(b_\delta\in(\mu_{11},\mu_{ij})\). Consider the same boundary-crossing event used in the proof of Proposition~\ref{prop: possible_limiting}, except that the initial sample mean of every other scenario \((i,\ell)\) is now based on \(n_0\) observations:
\begin{eqnarray*}
\mathcal E_{ij}^{\rm G}(b_\delta)
=
\left\{L_{ij}(b_\delta)=0\right\}
\cap\bigcap_{\ell\neq j}\left\{\bar X_{i\ell}(n_0)\leq b_\delta\right\}
\cap\bigcap_{\ell=1}^m\left\{U_{1\ell}(b_\delta)=0\right\}.
\end{eqnarray*}
On \(\mathcal E_{ij}^{\rm G}(b_\delta)\), the empirical worst-case mean of alternative 1 always lies below \(b_\delta\), while scenario \((i,j)\) remains strictly above \(b_\delta\). Every other scenario of alternative \(i\) begins below \(b_\delta\). It follows inductively that alternative \(i\) is never selected for an \(m\)-step and that scenario \((i,j)\) remains its empirical worst-case scenario. Hence, \(r_i^k(t)\to\infty\), and Assumption~\ref{assu:exploration_k} implies that scenario \((i,j)\) is sampled infinitely often. Every other scenario of alternative \(i\) retains its initial sample size \(n_0\).
Moreover, for every \(\ell\neq j\),
\[
\Pr\left\{\bar X_{i\ell}(n_0)\leq b_\delta\right\}
=\Phi\left(\frac{\sqrt{n_0}(b_\delta-\mu_{i\ell})}{\sigma_{i\ell}}\right)
=a_{i\ell}(n_0).
\]
Therefore, independence across scenarios and Equations~\eqref{eq:boundU} and~\eqref{eq:boundL} give
\begin{eqnarray*}
\notag &&\Pr\left\{\lim_{N\rightarrow\infty}n_{ij}(N)=\infty,\
\lim_{N\rightarrow\infty}n_{i\ell}(N)<\infty,\ \forall\,\ell\neq j\right\}  \geq\Pr\left\{\mathcal E_{ij}^{\rm G}(b_\delta)\right\}
\geq b_{ij}\prod_{\ell\neq j}a_{i\ell}(n_0)\prod_{\ell=1}^m b_{1\ell}>0.
\end{eqnarray*}
This proves Proposition~\ref{prop: possible_limiting_general}. If \(\mu_{ij}<\mu_{i1}\), then \(j\) is not a true worst-case scenario, and every scenario \(\ell\) satisfying \(\mu_{i\ell}=\mu_{i1}\) has \(\ell\neq j\). All such scenarios are sampled only finitely often on \(\mathcal E_{ij}^{\rm G}(b_\delta)\).  \hfill\Halmos
\end{proof}

\section{Sample Allocation Pattern of Two Existing DRR\&S Procedures}
\label{sec:existingsamplepath}
Here we provide some empirical evidence of the unexpected sample allocation behavior exhibited by the R-OCBA \citep{itemGao2017robust} and AR-OCBA procedures \citep{itemWan2024new}. We adopt the same experimental setup as in Sections~\ref{subsec:num_properties of AA} and~\ref{subsubsec: detail_GAA_exp}, using the MM-CV configuration, and setting a larger total sampling budget $N=(n_0+n_1)km$ with $n_1 = 150{,}000$. For each procedure, we visualize the final sample allocation across scenarios in Figure~\ref{fig:samplesizeOCBA} from two sample paths in which a correct selection is achieved. From the figure, we observe that although both R-OCBA and AR-OCBA are sequential heuristics, they do exhibit an additive pattern, concentrating sampling effort on $k + m - 1$ scenarios. However, this additive structure appears in an unexpected form: the set of $k + m - 1$ most sampled ``critical'' scenarios is not aligned with the hypothesized structure—namely, all scenarios of the best alternative and the $(i,1)$ worst-case scenario of each non-best alternative. Instead, the specific scenarios receiving the most observations vary across sample paths, and the worst-case scenario $(i,1)$ is usually \emph{not} among them. 
These observations conflict with the hypothesized  ``critical'' scenarios resulting from a static budget allocation analysis, thereby motivating a closer examination of additivity.
 
\begin{figure}[htbp]
    \centering
    \caption{Sample Allocation Pattern of R-OCBA and AR-OCBA when \(k = 10, m = 5\)}
    \includegraphics[width=0.9\linewidth]{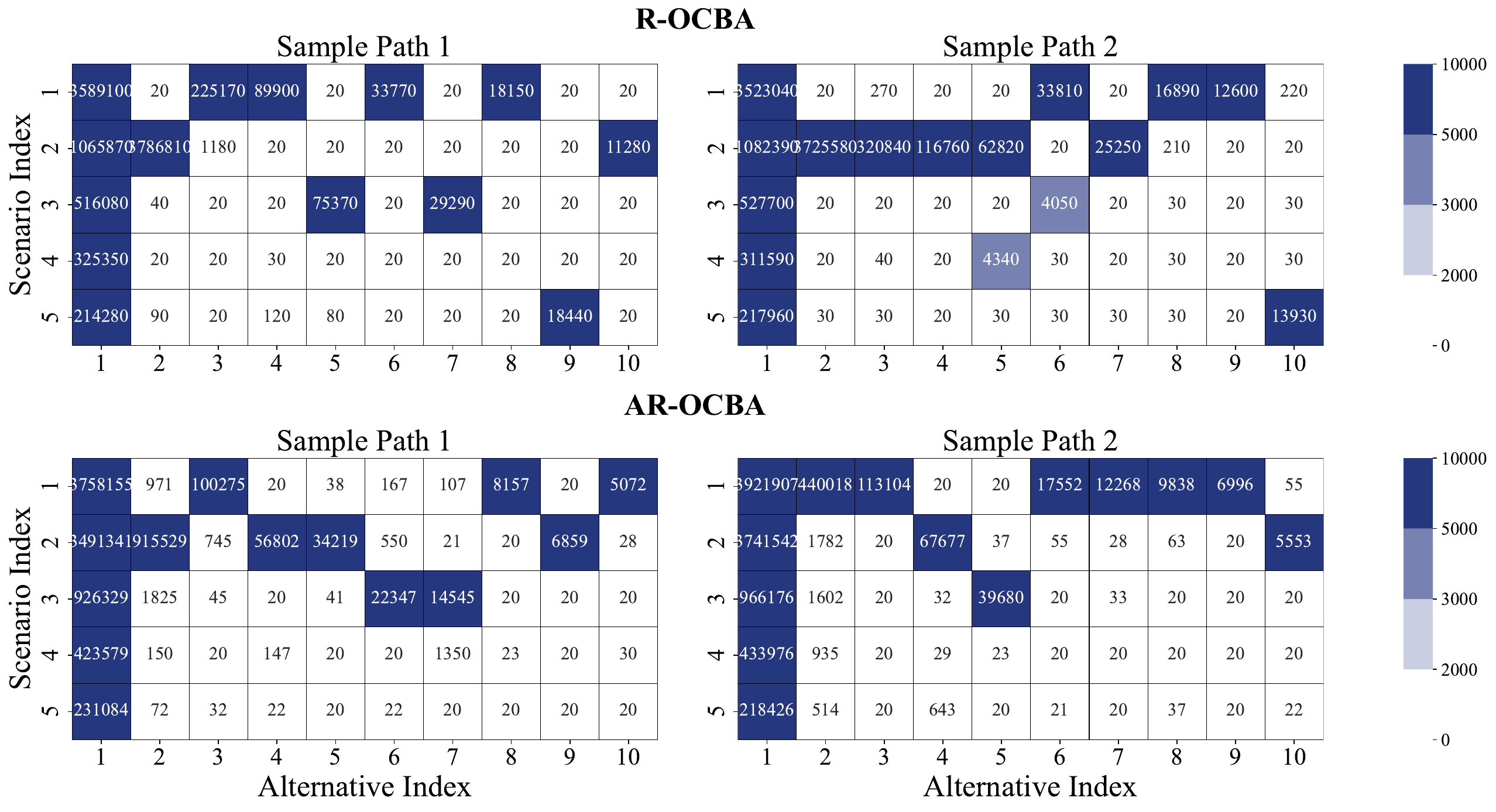}
        \label{fig:samplesizeOCBA}
\end{figure} 

\SingleSpacedXI

\end{APPENDICES} 

\end{document}